\pdfoutput=1

\documentclass[11pt]{article}

\usepackage[]{acl}

\usepackage{times}
\usepackage{latexsym}

\usepackage[T1]{fontenc}

\usepackage[utf8]{inputenc}

\usepackage{microtype}
\usepackage{algpseudocode}
\usepackage{cite}
\usepackage{amsmath,amssymb,amsfonts}
\usepackage{graphicx}
\usepackage{textcomp}
\usepackage{xcolor}
\usepackage{url}
\usepackage{listings}
\usepackage{helvet}
\usepackage{courier}
\usepackage{graphicx}
\usepackage{tabularx}
\usepackage{longtable}
\usepackage{booktabs}
\usepackage{caption}
\usepackage{amsmath}
\usepackage{subfigure}
\usepackage{hyperref}
\usepackage{xurl}
\usepackage{xspace} 
\usepackage{CJKutf8}
\usepackage{upgreek}
\usepackage{amsthm}
\usepackage{multirow}
\usepackage[ruled,linesnumbered]{algorithm2e}
\usepackage{pgfplots}
\pgfplotsset{compat=1.17} 
\usepackage{scalefnt} 
\newcommand{\eat}[1]{}

\newcommand{\laks}[1]{}

\newtheorem{theorem}{Theorem}

\definecolor{color1}{RGB}{141,20,80}
\definecolor{color2}{RGB}{34,68,122}
\definecolor{color3}{RGB}{32,99,23}
\definecolor{color4}{RGB}{123,89,4}
\definecolor{darkred}{rgb}{0.8, 0.0, 0.0}
\definecolor{darkblue}{rgb}{0.0, 0.0, 0.85}

\newcommand{\negative}[1]{\textcolor{darkred}{#1}}
\newcommand{\positive}[1]{\textcolor{darkblue}{#1}}

\title{DuNST: Dual Noisy Self Training for \\ Semi-Supervised Controllable Text Generation}

\author{Yuxi Feng\textsuperscript{1}, Xiaoyuan Yi\textsuperscript{2}\thanks{\text{ }\text{ }Work done during Yuxi Feng's internship at Microsoft Research Asia mentored by Xiaoyuan Yi.}, Xiting Wang\textsuperscript{2}, Laks V.S. Lakshmanan\textsuperscript{1}, Xing Xie\textsuperscript{2}\\ \textsuperscript{1}The University of British Columbia, Vancouver, Canada\\ 
\textsuperscript{2}Microsoft Research Asia, Beijing, China\\
\texttt{\{fyx14, laks\}@cs.ubc.ca,}\\
\texttt{\{xiaoyuanyi, xitwan, xing.xie\}@microsoft.com}}

\begin{document}
\maketitle

\begin{abstract}
Self-training (ST) has prospered again in language understanding by augmenting the fine-tuning of big pre-trained models when labeled data is insufficient. However, it remains challenging to incorporate ST into attribute-controllable language generation. Augmented only by self-generated pseudo text, generation models \emph{over-exploit} the previously learned text space and \emph{fail to explore} a larger one, suffering from a restricted generalization boundary and limited controllability. In this work, we propose DuNST, a novel ST framework to tackle these problems. DuNST jointly models text generation and classification as a dual process and further perturbs and escapes from the collapsed space by adding two kinds of flexible noise. In this way, our model could construct and utilize both pseudo text generated from given labels and pseudo labels predicted from available unlabeled text, which are gradually refined during the ST phase. Theoretically, we show that DuNST can be viewed as enhancing the exploration of the potentially larger real text space while maintaining exploitation, guaranteeing improved performance. Experiments on three controllable generation tasks show that DuNST significantly boosts control accuracy with comparable generation fluency and diversity against several strong baselines.
\end{abstract}

\section{Introduction}\label{sec:introduction}
\begin{figure}[t]
\center
\includegraphics[width=0.45 \textwidth]{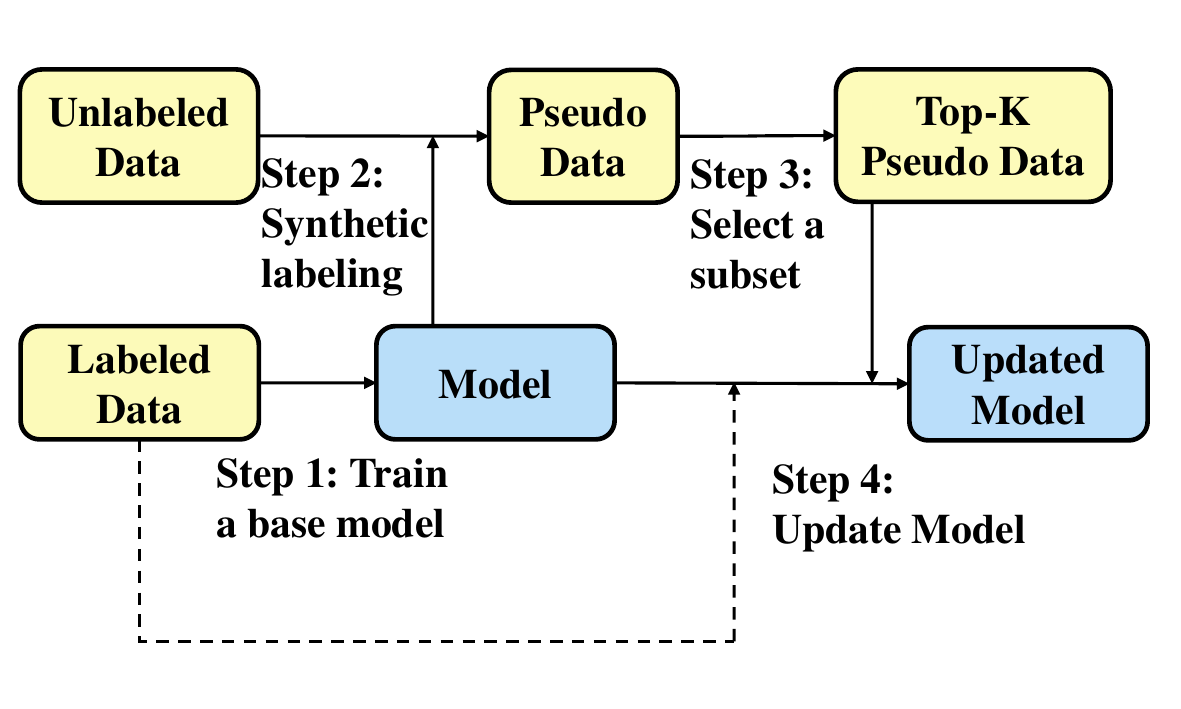}
\caption{Classic Self-training (ST) procedure. ST trains a base classifier on a small labeled dataset, then iteratively predicts pseudo labels for unlabeled data to augment the original labeled training set, and finally fits the model to the augmented dataset.} 
\label{fig_intro} 
\end{figure}

Recently, Pretrained Language Models (PLM)~\citep{Liu2019RoBERTaAR,dong2019unified,radford2019language,2020t5} have shown superiority in Natural Language Processing (NLP). However, the ever-growing size of these models demands more training data, which destabilizes the fine-tuning of PLMs when labeled data is highly insufficient~\citep{zhang2020revisiting}. In this case, \emph{Self-training (ST)}~\citep{scudder1965probability,yarowsky1995unsupervised,grandvalet2004semi}, a classical semi-supervised paradigm, has come to the fore again. As depicted in Fig.~\ref{fig_intro}, ST produces pseudo labels for text using a classifier and then retrains the classifier with augmented data in an iterative process. By this means, ST utilizes massive unlabeled text to denoise the pseudo-annotated neighbors and improve the generalization on real data~\citep{wei2020theoretical,zhang2022does}, boosting various Natural Language Understanding (NLU) tasks~\citep{mukherjee-awadallah-2020-ust,li-etal-2021-task-adaptive}.

Nevertheless, how to apply ST to Natural Language Generation (NLG), especially the data-hungry attribute-controllable NLG, remains an open question. Different from typical NLU tasks like text classification, controllable NLG takes an attribute label as input to generate a textual sequence meeting the given attribute rather than predicting labels given input text. 
This brings two new challenges for ST. \textbf{Challenge 1}: since model inputs become discrete labels, there is no massive unlabeled data for the NLG model to extend the learned distribution boundary. \textbf{Challenge 2}: augmented only by self-generated text, NLG models focus on exploitation and cannot explore a larger space. As a result, classic ST merely works for a few NLG tasks, \textit{e.g.}, Machine Translation~\citep{He2020Revisiting,jiao-etal-2021-self} where adequate in-domain text exists, but fails to benefit controllable NLG.

To handle these challenges, we propose a novel \textbf{Du}al \textbf{N}oisy \textbf{S}elf \textbf{T}raining (\textbf{DuNST}), for semi-supervised controllable NLG. DuNST jointly learns to generate text from given attribute labels and predict labels for text, characterizing these two directions as a dual variational generation process. Such duality allows our model to leverage not only generated pseudo text but also pseudo labels predicted for available unlabeled text. Both generation and classification would be augmented by the two kinds of pseudo data and thus gradually refined during the ST process, handling \emph{Challenge 1}. Besides, DuNST incorporates two new types of flexible noise into generated pseudo text, namely softmax temperature and soft pseudo text, to further perturb and escape from the text space learned at the previous ST iteration, which helps propagate local smoothness and enhance robustness~\citep{Xie2020SelfTrainingWN,chen-etal-2021-revisiting}, addressing \emph{Challenge 2}. Our method can be theoretically regarded as exploring a larger potential space,  thus facilitating an extended generalization boundary and improved attribute coverage, balancing exploration and exploitation better. Hence, DuNST could further boost controllability while maintaining comparable generation fluency and diversity. Our code is available at \url{https://github.com/peterfengyx/DuNST}.

In summary, our contributions are as follows:
\begin{itemize}
\item To the best of our knowledge, we are the first to incorporate Self-training into semi-supervised controllable language generation and propose a novel and effective ST method.

\item We demonstrate that DuNST explores a larger potential text space and extends the generalization boundary, providing a theoretical interpretation for our method.

\item We conduct thorough experiments on three attribute-controllable generation tasks and demonstrate the superiority of DuNST in improving control accuracy with competitive quality of the generated text, further extending the capacity of powerful PLMs for NLG.

\end{itemize}
\section{Related Work}
\paragraph{Controllable Language Generation:}
Attribute-controllable language generation aims to generate high-quality text satisfying desired attributes, \textit{e.g.}, sentiment, topic, and style, which could facilitate diverse downstream applications, such as stylistic writing~\citep{ficler-goldberg-2017-controlling,yi2020mixpoet} and language detoxification~\citep{gehman-etal-2020-realtoxicityprompts}. In the era of PLM, an effective paradigm for this task is fine-tuning PLMs on datasets containing labeled text~\citep{Keskar2019CTRLAC, gururangan-etal-2020-dont}. Nonetheless, with the increasing scale of PLMs, inadequate labeled data severely hampers the effectiveness of fine-tuning~\citep{zhang2020revisiting}.
As a remedy, two lines of research have been developed. \emph{Lightweight tuning} searches a trigger~\citep{sheng-etal-2020-towards} or optimizes only a few parameters~\citep{ribeiro-etal-2021-structural,yang2023uddia} or prefix~\citep{li-liang-2021-prefix,qian-etal-2022-controllable}, requiring much less training data. \emph{Plug-in control} steers the generation probability of NLG models towards target attributes through updating cached hidden states~\citep{Dathathri2020Plug} or rectifying the output distribution~\citep{liu-etal-2021-dexperts,krause2021gedi,yang-klein-2021-fudge} guided by off-the-shelf attribute classifiers or conditional PLMs at the inference time, without fine-tuning the generators. Despite no/weak dependence on labeled data, these two lines of work would cause limited control accuracy or decreased fluency.

\paragraph{Self-training:} Recently, Self-training has flourished again by iteratively generating pseudo labels and augmenting the tuning of data-hungry PLMs, showing great advantages in further enhancing NLU~~\citep{meng-etal-2020-text,vu-etal-2021-strata,du-etal-2021-self,bhat-etal-2021-self,chen-etal-2021-revisiting} and Neural Machine Translation (NMT)~\citep{He2020Revisiting, jiao-etal-2021-self} where massive unlabeled input text exists. Beyond naive ST, \citet{mukherjee-awadallah-2020-ust} select unlabeled instances based on the prediction uncertainty of the classifier to improve model confidence. \citet{jiao-etal-2021-self} also collect informative (high-uncertainty) monolingual sentences to enhance the translation quality of hard examples.  Relevant to our work, \citet{He2020Revisiting} corrupts the pseudo target sentences in NMT by synthetic noise like token shuffle or paraphrase to propagate local smoothness. However, as mentioned in Sec.\ref{sec:introduction}, due to \emph{Challenges 1\&2}, it is challenging to apply these ST methods to attribute-controllable NLG directly.

\paragraph{Dual Learning and Variational Generation:}  Dual Learning (DL)~\citep{he2016dual} has been traditionally proposed and applied in NMT and then extended to joint optimization of NLU-NLG tasks~\citep{xia2017dual}, which is promising for tackling Challenge 1. \citet{tseng-etal-2020-generative} successfully combined table-to-text and text-to-table generation, but their model cannot simultaneously optimize the two directions with shared features, not compatible with our design for Challenge 1. Variational Generation~\citep{kingma2014autoencoding} has proven to be effective in learning flexible semantic properties and thus generating more diverse and controllable text~\citep{hu2017toward,li-etal-2020-optimus,hu-etal-2022-fuse}, more suitable for our scenarios.


Unlike the aforementioned work, we revisit the challenges of incorporating Self-training with controllable generation and utilize the duality and flexible noise to handle these challenges, leading to a novel and practical ST framework.
\section{Methodology}

\subsection{Formulation and Overview}
Let $\mathbf{x}$ be the text, $y$ be the attribute label, $D_L\!=\!\{\mathbf{x}_i,y_i\}$ be a labeled dataset with paired text and its corresponding label, and $D_U\!=\!\{\mathbf{x}_i\}$ be an unlabeled dataset from the same domain. We aim to learn an attribute-controllable generator $\mathcal{G}\!=\!q_{\theta}(\mathbf{x}|y)$ parameterized by $\theta$ (\textit{e.g.}, a large PLM) to generate high-quality text $\mathbf{x} \sim q_{\theta}(\mathbf{x}|y)$ (in an auto-regressive manner) satisfying the given label $y$. We also endow our model with the ability to produce pseudo attribute labels for $\mathbf{x}\!\in\! D_U$ through jointly learning a text classifier $\mathcal{C}\!=\!q_{\phi}(y|\mathbf{x})$. We simultaneously model and optimize $\mathcal{G}$ and $\mathcal{C}$ with a shared PLM as a dual process (Sec.~\ref{subsec_dvae}). 

During the training of DuNST (Sec.~\ref{subsec_dunst}), the pseudo labels predicted by $\mathcal{C}$ help cover more unseen samples and hence extend the learned distribution boundary (\emph{tackling Challenge 1}), while the noisy pseudo text generated by $\mathcal{G}$ helps perturb the previously learned space, further improving generalization (\emph{addressing Challenge 2}). Though we emphasize generation in this work, both $\mathcal{G}$ and $\mathcal{C}$ would be promoted and thus keep refining the augmentation data during ST, which acts as a \emph{joint exploration and exploitation} process (Sec.\ref{subsec_theorem}).

\subsection{Dual Generation and Classification}
\label{subsec_dvae}
We jointly learn the conditional distribution of text $q_{\theta}(\mathbf{x}|y)$ and label $q_{\phi}(y|\mathbf{x})$ to match the real ones. However, we don't directly optimize them with traditional cross-entropy loss but resort to the variational approaches~\citep{kingma2014autoencoding}. In detail, we involve a latent variable $\mathbf{z}$ to capture the underlying semantics and hence have $q(\mathbf{x}|y) \!= \!\int q(\mathbf{x},\mathbf{z}|y)d\mathbf{z}$. We could sample a generated text $\mathbf{x}$ by the decomposition $q(\mathbf{x},\mathbf{z}|y)\!=\!q(\mathbf{x}|\mathbf{z},y)*q(\mathbf{z}|y)$. 
To this goal, we minimize a generation loss as:
\begin{align}
\mathcal{L}_g &= - \mathbb{E}_{p_{\psi}(\mathbf{z}|\mathbf{x},y)}[\log q_{\theta}(\mathbf{x}|\mathbf{z},y)] \notag \\
& + \text{KL}[p_{\psi}(\mathbf{z}|\mathbf{x},y)||q_{\theta}(\mathbf{z}|y)], \label{vaegen}
\end{align}
where 
$p_{\psi}(\mathbf{z}|\mathbf{x},y)$ and $q_{\theta}(\mathbf{z}|y)$ are approximated posterior and prior distributions of $\mathbf{z}$ and $\text{KL}$ is the Kullback–Leibler divergence, respectively. Optimizing this loss is equivalent to maximizing a lower bound of $q_{\theta}(\mathbf{x}|y)$.

The posterior $p_{\psi}(\mathbf{z}|\mathbf{x},y)$ is typically assumed as a multivariate Gaussian $\mathbb{N}(\mu_{post},\sigma_{post})$ and approximated by $[\mu_{post},\log\sigma_{post}]\!=\!\text{MLP}([\mathbf{h}_{\mathbf{x}},\mathbf{h}_y])$ with $\mathbf{h}_{\mathbf{x}}\!=\!\text{Encoder}(\mathbf{x})$, where $\mathbf{h}_{y}$ is the label embedding of $y$. $\text{Encoder}$ is a Transformer~\citep{vaswani2017attention} encoder, and MLP is a multilayer perceptron. Similarly, we could build the prior $q_{\theta}(\mathbf{z}|y)\sim\mathbb{N}(\mu_{gen_prior},\sigma_{gen_prior})$ where $[\mu_{gen_prior},\log\sigma_{gen_prior}]\!=\!\text{MLP}(\mathbf{h}_y)$.

Symmetrically, we optimize classification by:
\begin{align}
    \mathcal{L}_c &= -\mathbb{E}_{p_{\psi}(\mathbf{z}|\mathbf{x},y)}[\log q_{\phi}(y|\mathbf{z},\mathbf{x})] \notag \\
    & + \text{KL}[p_{\psi}(\mathbf{z}|\mathbf{x},y)||q_{\phi}(\mathbf{z}|\mathbf{x})].
\end{align}

The text is generated by an autoregressive Transformer decoder $\mathbf{x}\!=\!
\text{Decoder}(\mathbf{z})$ and the label is predicted by $y\!=\!\text{MLP}(\mathbf{z})$ with $\mathbf{z}$ drawn from the posterior distribution in training and from the prior ones in testing. $\mathcal{G}$ and $\mathcal{C}$ share most parameters (\textit{e.g.}, encoder), as well as the same posterior distribution $p_{\psi}(\mathbf{z}|\mathbf{x},y)$, to enhance the connection of text and corresponding labels, and better utilize the knowledge learned via the two directions.

The final loss is computed as follows:
\begin{equation}
\label{dvaeloss}
\begin{aligned}
    \mathcal{L} =& \lambda_g \mathcal{L}_g + \lambda_c \mathcal{L}_c,
\end{aligned}
\end{equation}
where $\lambda_g$ and $\lambda_c$ are hyper-parameters to balance the importance of classification and generation. We will show later that such variational dual learning further boosts controllability and text diversity (Sec.~\ref{subsec_ablation}) and helps refine pseudo labels (Sec.~\ref{subsec_analysis}).

\subsection{Dual Noisy Self-training}
\label{subsec_dunst}
As discussed in Sec.~\ref{sec:introduction}, augmented only by self-generated text, the model would increasingly enhance the exploitation of the previously learned space but fail to explore more, resulting in constrained attribute distributions and thus marginal improvement of control accuracy (\emph{Challenge 2}, see Table \ref{tab:mainresult}). Injecting noise into pseudo text is a practical way to facilitate exploration. However, the typical synthetic noise~\citep{He2020Revisiting} (\textit{e.g.}, randomly shuffle tokens in pseudo text) encourages isotropic exploration, which may diverge far from the valid space and get too noisy for NLG.
\begin{algorithm}[tp]
 \caption{Training Process of DuNST}
 \label{alg:dunst}
\KwIn{Labeled set $D_L$, unlabeled set $D_U$,  attribute set $Y$.}
\BlankLine
Jointly train base model $\mathcal{G}$, $\mathcal{C}$ on $D_L$ by optimizing Eq.\eqref{dvaeloss}, store the best $\mathcal{G}_0$, $\mathcal{C}_0$.

\For{$epoch\leftarrow 1$ \KwTo MaxEpoch}{

\For{$\mathbf{x}_i$ \textbf{in} $D_U$}{
$\hat{y}_i=\mathcal{C}_{epoch-1}(\mathbf{x}_i)$
}
Build pseudo label set: $D_{PL} \!=\! \{\mathbf{x}_i,\hat{y}_i\}$

\For {$y_j$ \textbf{in} $Y$}  {

Sample $t$ priors: $\{z_k\}_{k=1}^t \sim q_{\theta}(\mathbf{z}|y_j)$

\For {k$\leftarrow$ 0 \textbf{to} t }{

\For {m $\leftarrow$ 0 \textbf{to} MaxLength }{
Compute soft pseudo token $\mathbf{d}_{k}^m$ using $\mathcal{G}_{epoch-1}$ and Eq.\eqref{softtoken}.
Set $y_k\leftarrow y_j$.
}
}
}
Build soft pseudo text: $D_{PT}\!=\!\{\mathbf{d}_k,y_k\}$

Train $\mathcal{G}_{epoch-1}$, $\mathcal{C}_{epoch-1}$ on $\{D_{PT},D_{PL}, D_L\}$ by optimizing Eq.\eqref{dvaeloss} and Eq.\eqref{sptloss}, update the parameters to $\mathcal{G}_{epoch}$ and $\mathcal{C}_{epoch}$.
}
\end{algorithm}

To address this problem, we propose two novel and effective types of soft noise to enable safer exploration, namely \emph{High-temperature Generation} and \emph{Soft Pseudo Text}, in what follows.

\paragraph{High-temperature Generation (HTG):}
We introduce temperature $\tau$ in the softmax layer:
\begin{equation}
\mathbf{d}^m=\sigma(\mathcal{G}(y,\mathbf{\hat{x}}_{<m},\mathbf{z})/\tau),
\label{softtoken}
\end{equation}
where $\mathbf{d}^m$ is the output token distribution for the $m$-th token, $\mathbf{\hat{x}}_{<m}$ is the previously generated $m-1$ tokens and $\sigma$ means softmax. Lower $\tau$ (\textit{e.g.}, $\tau\!<\!1$ ) leads to a sharper distribution and thus motivates more certain output (usually used in NMT). Differently, we choose $\tau\!>\!1$ to encourage more diverse but semantically reasonable (high generation probability) tokens which could enhance local smoothness and help explore more potential directions. Besides, the degree of noise is easy to control by adjusting $\tau$ for a better trade-off.

\paragraph{Soft Pseudo Text (SPT):} HTG improves the diversity of pseudo text, but also takes the risk of sampling invalid tokens and propagating errors in an autoregressive generation. Moreover, HTG produces discrete pseudo text (a point in text space) and thus requires numerous sampled pseudo text (points) to cover a small neighborhood (Fig.~\ref{fig_theorem}). Therefore, we further propose to generate soft pseudo text, where we directly store the output token distribution $\mathbf{d}$ and let $\mathcal{G}$  directly learn to reproduce $\mathbf{d}$. Then we replace Eq.\eqref{vaegen} with:
\begin{equation}
\mathcal{L}_g^{'}=\left\{
\begin{aligned}
    &-\log q_{\theta}(\mathbf{x}|\mathbf{z},y)+\\ &\text{KL}[p_{\psi}(\mathbf{z}|\mathbf{x}, y)||q_{\theta}(\mathbf{z}|y)],  \mathbf{x},y\in D_L, D_{PL}\\
    &\text{KL}[\mathbf{d}||q_{\theta}(\mathbf{x}|\mathbf{z},y)]+\\ &\text{KL}[p(_{\psi}\mathbf{z}|\mathbf{x}, y)||q_{\theta}(\mathbf{z}|y)], \mathbf{x},y\in D_{PT}.
\end{aligned}
\label{sptloss}
\right.
\end{equation}
Such SPT acts as a kind of Knowledge Distilling~\citep{hinton2015distilling} in an iterative manner. In this way, we avoid losing relevant semantic information in $\mathbf{d}$ and reduce needed samples, further extending the generalization boundary (see Table~\ref{tab:ablation}).

The complete algorithm is described in Alg.~\ref{alg:dunst}.

\subsection{Theoretical Analysis}
\label{subsec_theorem}
To understand why DuNST could work well, we interpret its advantages with the following theorem:
\begin{figure}[tbp]
\center
\includegraphics[width=0.46 \textwidth]{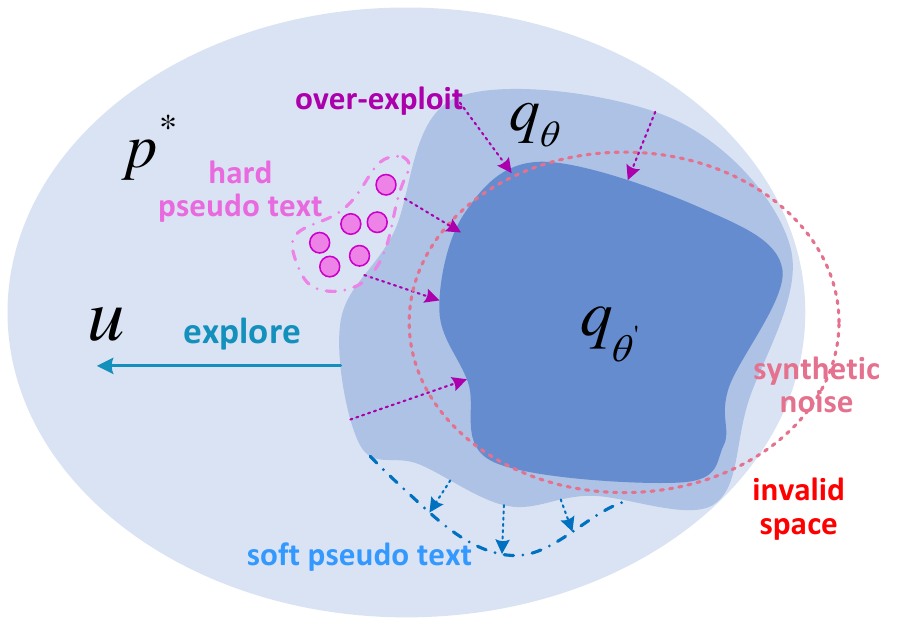}
\caption{The illustration of Theorem 1.} 
\label{fig_theorem} 
\end{figure}

\begin{theorem}
\label{thrm}
Optimizing the training objective of DuNST is equivalent to approximately minimizing the upper bound of 
\begin{equation}
\label{eqkl}
\begin{aligned}
    \text{KL}\left[p^{*}||q_\theta\right] 
    + \text{KL}\left[p_{\theta'}||q_\theta\right] 
    +\text{KL}\left[u||q_\theta\right],
\end{aligned}
\end{equation}
where $p^{*}$ is the real text distribution, $q_\theta$ and $q_{\theta'}$ are models estimated at the current and last ST iteration, respectively, and $u$ is a noise distribution.
\end{theorem}
\emph{Proof}. See Appendix~\ref{sec:apdix-theorem}.

In Theorem \ref{thrm}, the first KL term corresponds to the optimization of Eq.(\ref{dvaeloss}) that approximates the real distribution. The second term corresponds to classic Self-training, which works as a regularization. As depicted in Fig.~\ref{fig_theorem}, such regularization forces the model to fit the already learned space, causing over-exploitation. The last one is the noise to enhance exploration. Compared to the isotropic synthetic noise (too noisy) and the hard pseudo text (too sparse), DuNST with soft pseudo text could explore potential directions, cover larger space more smoothly, and thus further push the boundary.
\begin{table*}[htbp]
\small
 \centering
 \begin{tabular}{lcccccccccl}\toprule
    & \multicolumn{5}{c}{Sentiment} & \multicolumn{5}{c}{Topic}
    \\
    \cmidrule(lr){2-6}\cmidrule(lr){7-11}
             & O-PPL $\downarrow$  & M-PPL $\downarrow$ & F1 $\uparrow$ & Dist $\uparrow$ &S-BL $\downarrow$  & O-PPL $\downarrow$  & M-PPL $\downarrow$ & F1 $\uparrow$ & Dist $\uparrow$ &S-BL $\downarrow$ \\\midrule
    Ground-Truth & 25.14&$-$ & 96.20 & 48.27 & 43.34 & 31.04&$-$ & 94.89 & 67.24 & 23.31 \\
    GPT2(raw) & 13.20&38.39&68.50&35.91&58.79 & 16.94 & 74.41 & 52.17 & 46.88&45.55\\
    \midrule
    \multicolumn{7}{l}{\textit{Finetuned PLM}} \\
    GPT2    &16.40&44.02&80.44&26.34&71.00 &22.22 &23.46&83.08&54.93&39.93\\
    UniLM & 25.20&54.33&75.35&31.05&66.97&55.79 &36.28&87.70&54.76&43.77 \\
    T5 & 25.69&34.97&83.77&30.03&69.57
    & 48.33&32.12&88.43&58.06&37.01\\
    \midrule
    \multicolumn{7}{l}{\textit{Lightweight method}}\\
    PF  &13.02&37.09&75.05&29.48&65.10
    &\textbf{20.27}&32.35&68.44&59.17&32.73\\
    Ctr-PF &13.01&37.12&77.33&29.63&\textbf{64.83}
    &20.41 &33.90&83.21&\textbf{60.34}&\textbf{31.20}\\
    \midrule
    \multicolumn{7}{l}{\textit{Self-Training with PLM}}\\
    PT    & 14.62 & 68.04&79.57&30.58&65.22 & 57.40&40.95&86.36&52.35&46.41\\
    PT(noise) &11.91 &44.31&77.46&25.40&72.19 &58.59&45.32&85.27&53.35&46.57 \\
    PT(noise)+PL &11.26&\textbf{33.85}&88.47&27.26&70.90 &32.36&\textbf{16.64}&89.70&53.79&47.95 \\
    PT(select)+PL &\textbf{10.89}&33.89&88.75&27.17&71.41 &33.23&16.66&90.52&53.71&47.69 \\
    \midrule
    DuNST  & 21.67&42.82&\textbf{93.05}&\textbf{31.79}&65.80
    & 34.73&33.58&\textbf{93.59}&59.42&37.02\\
    \bottomrule
 \end{tabular}
 \caption{Results on IMDb dataset (sentiment) and AGNews dataset (topic).}
 \label{tab:mainresult}
\end{table*}
\section{Experiments}
\label{sec:experiment}
\subsection{Tasks}
We conduct exhaustive experiments on three controllable generation tasks, described below: 

\paragraph{Sentiment control with prompt:} We evaluate the controllability of sentiment on the IMDb movie review dataset \citep{maas-etal-2011-learning}. Following~\citep{Dathathri2020Plug}, we use their 15 manually created prompts and another 85 sampled from IMDb (100 in total) as model input and generate 10 samples for each prompt and each sentiment.

\paragraph{Topic control w/o prompt:} We use the AGNews dataset ~\citep{Zhang2015CharacterlevelCN} to evaluate topic controllability. We assess our model's ability to generate from scratch on this dataset and sample 300 generations for each topic.

\paragraph{Text detoxification:} We use the Jigsaw Toxic Classification Dataset. Following~\citep{qian-etal-2022-controllable}, we use the 203 ``challenging'' prompts (toxicity < 0.5) from~\citep{gehman-etal-2020-realtoxicityprompts}, and generate 10 non-toxic sentences for each prompt.

We sample 5\% of IMDb training samples as labeled data and directly take their provided unlabeled set. Since there is no separate unlabeled text in AGNews, we sample 3\% of training samples as labeled data and use the others as unlabeled ones. For a fair comparison, we keep the ratio of the original human-labeled data/generated pseudo text/unlabeled data with pseudo label ($D_L$/$D_{PT}$/$D_{PL}$ (in Alg.~\ref{alg:dunst})) to 1:1:30. More details of the dataset are provided in Appendix \ref{sec:datasetdesc}.
\subsection{Experimental Settings}
We use UniLM-base-cased \citep{dong2019unified} as the shared encoder and decoder of DuNST. The dimension of latent $z$ is 128 for sentiment control and 256 for topic control. We use AdamW~\citep{Loshchilov2019DecoupledWD} with learning rate $=$ 5e-5, warm-up steps $=$ one epoch, and batch size $=$ 8 for optimization across all tasks. $\lambda_g\!=\!1$ and $\tau\!=\!5$ for all tasks, $\lambda_c$ is 10 for IMDb and 1 for AGNews. As a common practice~\citep{holtzman2019curious}, we use top-$p$ with $p\!=\!0.9$ sampling method for decoding. To stabilize training, we further incorporate BOW~\citep{wang2017diverse} and annealing~\citep{fu-etal-2019-cyclical} techniques. Following ST in NLU~\citep{mukherjee-awadallah-2020-ust}, we start ST from a base model tuned on $D_L$ without any sample selection as in~\citep{vu-etal-2021-strata}. More implementation details are provided in Appendix \ref{sec:impdetail}.

\subsection{Evaluation Metrics}
We mainly focus on the controllable NLG side, considering the following four kinds of metrics. Those for classification are provided in Appendix~\ref{sec:apdix-toxic}.

\paragraph{Fluency:} We evaluate generation fluency by the perplexity (PPL) of generated text measured by GPT2-XL~\citep{radford2019language}, \textit{i.e.}, \textbf{Output PPL}.

\paragraph{Generalizability:} We calculate the PPL of each model on the held-out test set in each dataset, \textit{i.e.} \textbf{Model PPL}, to evaluate how well the model could generalize to test data in a specific unseen domain.

\paragraph{Controllability:} We evaluate the control accuracy through classification Macro-F1(\textbf{F1})) on the generated text by RoBERTa-large based classifiers fine-tuned on corresponding full training data for sentiment and topic, respectively. For toxicity evaluation, we use the  Perspective API\footnote{\url{https://www.perspectiveapi.com/}}.

\paragraph{Diversity:} To evaluate the diversity of generated text, we consider \textbf{Dist-n}~\citep{li-etal-2016-diversity} and \textbf{Self-BLEU (S-BL)}~\citep{zhu2018texygen}. 

More metrics details are described in Appendix \ref{sec:apdix-metric}.

\subsection{Baselines}
We compare our model with three kinds of (supervised or semi-supervised) strong NLG baselines.

\paragraph{Finetune PLM:} We finetune different powerful PLMs on each downstream dataset, including GPT2~\citep{radford2019language}, UniLM~\citep{dong2019unified} and T5~\citep{2020t5}.

\paragraph{Lightweight fine-tuning methods:} 
(1) Prefix-tuning (PF)~\citep{li-liang-2021-prefix}: this method only tunes the prefix and freezes all parameters of the PLM, requiring less data. (2)
Ctr-PF\citep{qian-etal-2022-controllable}: A contrastive version of PF.

\paragraph{Self-training methods:}
(1) PT: the classical Self-training~\citep{grandvalet2004semi}, which generates pseudo text in each ST iteration and updates parameters with both real and pseudo text from the last iteration.
(2) PT(noise): Noisy Self-training~\citep{He2020Revisiting}, which brings synthetic noise (token drop, swap and mask) to the pseudo text for self-training.
(3) PT(noise)+PL: We combine PT(noise) and \emph{pseudo labeling} to produce and utilize both pseudo text and pseudo labels, which are predicted from the real unlabeled text by a BERT-base~\citep{devlin-etal-2019-bert} fine-tuned on our labeled data.
(4) PT(select)+PL: PT(select) is a modified ST method with sample selection~\citep{mukherjee-awadallah-2020-ust}, which over-generates noisy pseudo text and selects high-quality ones by the classifier confidence and uncertainty. 

For a fair comparison, we choose the PLM with the best fine-tuning performance on each task (and a similar model size to ours) as the backbone of these ST variants (GPT2 for sentiment and UniLM for the others). Besides, we also provide the evaluation results of Ground Truth as an upper bound of performance. We give more details of the baseline models above in Appendix \ref{sec:apdix-baseline}.

\subsection{Results}
As shown in Table \ref{tab:mainresult}, on both tasks, our DuNST achieves significant improvement in controllability compared to fine-tuned PLMs and lightweight tuning and is comparable in fluency, generalizability, and diversity. Fine-tuned PLMs obtain limited F1 improvement but severely decreased diversity (+6.7 S-BLEU at most), indicating they are overfitted to these few labeled data points and fail to cover larger attribute spaces. PF and Ctr-PF only reduce required data but perform even worse than tuned PLMs. The unnatural O-PPL (much lower than that of ground truth) shows they lose the capacity of PLMs and cause degenerated results. In contrast, thanks to the duality, DuNST simultaneously refines pseudo labels and enhances the quality and diversity of pseudo text in an iterative manner, boosting controllability and diversity (\emph{Challenge 1}).

We also have some interesting findings about existing self-training methods. 1) The classic ST method even hurts controllability and generalizability in the sense of \emph{Challenge 2}. As discussed in Sec.~\ref{sec:introduction}, merely self-generated text over-stresses exploitation of the learned space and hinders exploration. 2) Traditional synthetic noise PT(noise) motivates isotropic exploration, which diverges from valid attribute distributions (poorer O/M-PPL). 3) Sample selection brings a marginal improvement but costs 50\% more training time. Thus we did not apply such a selection in DuNST. 4) Additional pseudo-labels significantly improve performance. However, unlike our dual method, the fixed pseudo labels by PT(noise)+PL cannot evolve during ST. By comparison, DuNST utilizes high-temperature sampling and soft text to introduce flexible noise, encouraging safer exploration and better controllability and diversity while maintaining good quality.

\begin{table}[tb]
\small
 \centering
 \begin{tabular}{crlll}\toprule
       & Model & Fluency $\uparrow$  & Novelty $\uparrow$ & Rel. $\uparrow$\\\midrule
   \multirow{3}{*}{\emph{Sentiment}}  & Ctr-PF & 3.23$^{**}$ & 3.38$^{**}$& 3.37$^{**}$\\
    & ST & 3.35 &3.65 & 3.83$^{**}$\\
    & DuNST  & \textbf{3.51} & \textbf{3.69} & \textbf{4.13} \\
    \midrule
    \multirow{3}{*}{\emph{Topic}} & Ctr-PF &3.66$^{**}$ & 4.16$^{**}$& 4.51$^{*}$ \\
    & ST &3.97 &4.43 & 4.57$^{*}$\\
    & DuNST  & \textbf{4.01}&\textbf{4.50}&\textbf{4.71}\\
    \bottomrule
 \end{tabular}
 \caption{Human evaluation results on sentiment/topic-controlled generation. ST refers to the best ST variant under automatic evaluation. We conduct Student t-test for statistical significance. Notation: $^{**}$: $p$-value$<0.01$, $^{*}$: $p$-value$<0.05$. The Cohen’s kappa score is 0.63, indicating a satisfactory inter-annotator agreement. }
 \label{tab:human}
\end{table}

Due to space limitations, we report the results of text detoxification under both automatic and human evaluation in Appendix \ref{sec:apdix-toxic}.

\subsection{Human Evaluation}
To better verify the effectiveness of DuNST, we also conduct a human evaluation. We generated 100 samples for each model and each task and invite 5 competent annotators to score the samples on \textbf{Fluency}, \textbf{Novelty}, and \textbf{Attribute Relevance}. As shown in Table \ref{tab:human}, DuNST consistently outperforms baselines on all three  criteria, which indicates that DuNST not only has better attribute controllability but also generates fluent and diverse texts. See Appendix \ref{sec:apdix-human} for detailed evaluation protocol.

\subsection{Ablation Study}
\label{subsec_ablation}
\begin{table}[htbp]
\small
 \centering
 \begin{tabular}{rcccl}\toprule
    & \multicolumn{4}{c}{IMDb} 
    \\
    \cmidrule(lr){2-5}
             & O-PPL $\downarrow$  &M-PPL $\downarrow$ & F1 $\uparrow$ &S-BL $\downarrow$  \\\midrule
    DuNST  & 21.67&42.82&\textbf{93.05}&\textbf{65.80}\\
    $-$Variational &19.67&38.56&92.12&66.21\\
    $-$SPT &\textbf{18.53}&\textbf{36.53}&91.64&67.07\\
    $-$PT&20.91&41.14&91.83&66.27\\
    $-$PL &47.45&197.27&83.41&66.17\\
    $-$PL$-$SPT& 48.56&219.30&80.85&66.12\\
    $-$PL$-$PT&42.12&147.14&82.89&68.91\\
    \bottomrule
 \end{tabular}
 \caption{Ablation study on IMDb dataset. PT: pseudo text. SPT: soft pseudo text. PL: pseudo label. The symbol $-$ means removing settings from DuNST. $-$Variational to jointly trained classifier and generator. $-$PL$-$PT reduces to naive dual variational learning.}
 \label{tab:ablation}
\end{table}
We conduct an ablation study on the IMDb dataset. As shown in Table \ref{tab:ablation}, we can find: 1) variational learning further enhances control accuracy and diversity with slight PPL loss, which is worthwhile since the generated text is already fluent enough (close to ground truth PPL). 2) pseudo labels lead to a significant improvement. 3) soft pseudo text outperforms the hard one on controllability and diversity but with marginal fluency loss. Solely hard pseudo text in ST limits model coverage, while the soft one brings a smoother noise and helps push the learned boundary.

\subsection{Analysis}
\label{subsec_analysis}
\begin{figure}[tb] 
\centering 
\resizebox{0.9\columnwidth}{!}{  
			\begin{tikzpicture} 
			\scalefont{0.8} 
			\begin{axis}[
			sharp plot, 
			xmode=normal,
			xlabel=Epoch, 
			ylabel=F1, 
			width=7.5cm, height=5cm,  
			xmin=0,xmax=26,  
			ymin=70, ymax=95,  
			xtick={0,5,10,15,20,25}, 
			ytick={70,75,80,85,90,95}, 
			xlabel near ticks, 
			ylabel near ticks, 
                ylabel style={rotate=-90},
			ymajorgrids=true, 
			grid style=dashed, 
			legend style={at={(0.7,0.02)},anchor=south}, 
			]
			
			\addplot+[very thick,mark=square,mark options={scale=0.7}, color=color1] plot coordinates { 
				(1,73.36)
				(2,83.07)
				(3,86.64)
				(5,88.05)
				(10,87.63)
                (15,91.00)
                (20,92.49)
                (25,93.59)
			};
			\addlegendentry{DuNST}
   
			\addplot+[very thick,mark=star,mark options={scale=1.4}, color=color2] plot coordinates {
                    (1,73.9)
				(2,83.18)
				(3,85.59)
				(5,90.33)
				(10,87.73)
                    (15,88.47)
                    (20,89.27)
                    (25,88.49)
			};
			\addlegendentry{$-$Dual} 

            \addplot+[dashed,very thick,mark=star,mark options={scale=0.8}, color=color3] plot coordinates {
                (0,86.0)
				(25,86.0)
			};
			\addlegendentry{Base} 
   
            \addplot+[dashed,very thick, mark options={scale=0.7}, color=color1] plot coordinates {
                (0,86.0) 
				(1,86.9)
				(2,87.6)
				(3,87.7)
				(5,87.8)
				(10,87.8)
                (15,88.3)
                (20,88.5)
                (25,88.6)
			};
   
			
            \addplot+[dashed,very thick,mark=square,mark options={scale=0.6}, color=color2] plot coordinates {
                (0,86.0)
                (1,74.4)
				(2,54.8)
				(3,50.0)
				(5,48.1)
				(10,44.8)
			};

			\end{axis}
			\end{tikzpicture}
		}

		\caption{F1 score over the number of training epochs on topic. Solid lines indicate generation controllability, while dashed ones refers to classification. The green line is classification F1 of our base model at epoch 0.} 
		\label{fig:dual}  
\end{figure}
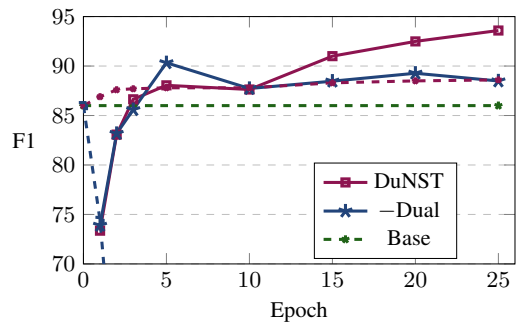
\paragraph{Effect of Duality:}
We compare our model with a variant (\textbf{$-$Dual}) where we annotate pseudo labels in advance and cut off classification losses through self-training. As depicted in Fig.~\ref{fig:dual}, since classification and generation share parameters, without optimizing the classifier and pseudo labels, the learned distribution $q_{\theta}$ would gradually shift and thus the classification performance greatly drops. As a result, generation F1 reaches its maximum soon and stops increasing. On the other hand, thanks to the simultaneously optimized classifier, DuNST keeps improving classification and refining pseudo labels, further enhancing controllability. 

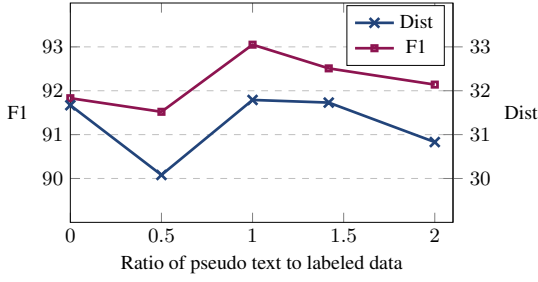
\begin{figure}[tbp] 

\centering 
\resizebox{0.95\columnwidth}{!}{  
			\begin{tikzpicture} 
			\scalefont{0.8} 
			\begin{axis}[
			sharp plot, 
			xmode=normal,
			xlabel=Ratio of pseudo text to labeled data, 
			ylabel=F1, 
			width=7.5cm, height=5cm,  
			xmin=0,xmax=2.1,  
			ymin=89, ymax=94,  
			xtick={0,0.5,1,1.5,2.0}, 
			ytick={90,91,92,93}, 
			xlabel near ticks, 
			ylabel near ticks, 
                ylabel style={rotate=-90},
			ymajorgrids=true, 
			grid style=dashed, 
			legend style={at={(0.9,1.1)},anchor=south}, 
			]
			\addplot+[very thick,mark=square,mark options={scale=0.6}, color=color1] plot coordinates {
                    (0,91.83)
                    (0.5,91.52)
                    (1,93.05)
                    (1.418,92.51)
                    (2, 92.14)
                    
			};
            \label{numpda}
			\end{axis}
   
    \begin{axis}[
    axis y line*=right,
    axis x line=none,
    xlabel=x, 
    ylabel=Dist, 
    width=7.5cm, height=5cm,
    xmin=0,xmax=2.1,
    ymin=29, ymax=34,
    ytick={30,31,32,33},
    tick align=outside, 
    ylabel style={rotate=-90},
    legend style={at={(0.85,0.99)},anchor=north}
    ]

\addplot[very thick,mark=x,color=color2,
mark options={scale=1.5}] plot coordinates {
    (2, 30.83)
    (1.418,31.73)
    (1,31.79)
    (0.5,30.08)
    (0,31.67)
};

\addlegendentry{Dist}
\addlegendimage{/pgfplots/refstyle=numpda}\addlegendentry{F1}

\end{axis}
\end{tikzpicture}
		}

		\caption{Generation F1 and diversity (Dist) with different numbers of pseudo text on IMDb dataset.} 
		\label{fig:numpt}  
\end{figure}
\paragraph{Number of pseudo text:}
We also evaluate DuNST on varying numbers of pseudo text, keeping other settings unaltered. As shown in Fig.~\ref{fig:numpt}, DuNST performs the best with equal size of pseudo text and labeled data. More pseudo text brings too much noise, which hurts generation quality as the model learns more noise than semantics. Too less pseudo text makes the model lose exploration ability and thus fail to extend the learned distribution boundary. Therefore, we should find a suitable noise level to balance exploration and exploitation.

\paragraph{Number of labeled data:}
Besides, we also assess our model with \emph{varying numbers of labeled training instances} (with the same unlabeled data size) and observe consistent superiority to baseline models. Though all models benefit from more annotations, our DuNST could quickly learn from pseudo labels and text and thus achieve better performance with much less labeled data. Detailed analyses are described in Appendix \ref{sec:apdix-analysis}. 

\input{fig_temp.tex}
\paragraph{Effect of Noise (temperature):}
To illustrate why noise encourages exploration and improves control, we plot the posterior of generations in different temperatures and visualize the estimated decision boundaries based on training data in Fig.~\ref{fig:zpost}. We find that higher noise leads to more challenging points (and thus more informative pseudo text) closer to the current boundary. Such refined pseudo data enables the model to learn to distinguish representations under different attributes better and push the generalization boundary, thus potentially improving generation controllability. Besides, the noisy pseudo text also helps further improve exploration and attribute coverage. We discuss it in Appendix \ref{sec:apdix-analysis}. 

Fig.~\ref{fig:temperature} shows the generation performance of DuNST
with different temperatures. We achieve the best controllability and diversity when the temperature is 5 in our setting. Less noise would lose the exploration ability and damage generalization, while too-noisy pseudo text runs a risk of approaching invalid space (Sec.~\ref{subsec_theorem}), which indicates a suitable temperature is necessary to balance exploration and exploitation.

\paragraph{Case Study:}
In order to verify the generation quality and attribute relevance, we present some cases sampled from different models in Table~\ref{tab:case}. We can see that Ctr-PF and ST (GPT2) suffer from repeating expressions (e.g., ``poor quality'' and ``got bored''). In contrast, DuNST produces more diverse and fluent expressions, which are also more faithful to the given negative attribute. We provide more generated cases in Appendix \ref{sec:apdix-case}.

\begin{table}[tbp]
 \centering
 \small
 \begin{tabular}{p{1cm}|p{6cm}}\toprule
         Model & Generation\\\midrule
    Ctr-PF & \textit{At 20 years old}, \negative{ichorous} and \negative{unrepentant}, ichanarito is a \positive{well-liked} celebrity, but a \negative{poor quality actor}, and a \negative{poor quality movie}... \\
    ST-(GPT2) & \textit{At 20 years old},  I grew up with it and \negative{had no expectations}.  I was a \positive{big fan} of the series, I thought it was \positive{great}, and I wanted to see more of it. But soon enough, I \negative{got bored} and just \negative{got bored} watching it.\\
    DuNST  & \textit{At 20 years old}, the \negative{poorest} hour of my life had me scratching my head. This was an \negative{insult} to the intelligence of mankind, and a complete \negative{disregard} for anyone else's intelligence. \negative{Do not watch} this movie. It is a \negative{disgrace} to any religion.\\
    \bottomrule
 \end{tabular}
 \caption{Sampled generations with specified negative sentiment and input prompt \textit{At 20 years old}. Words in \positive{blue}/\negative{red} are positive/negative indicators, respectively.}
 \label{tab:case}
\end{table}

\section{Conclusion and Future Work}
We propose a novel DuNST method to apply Self-training to semi-supervised controllable NLG. DuNST (1) jointly optimizes generation and classification via a dual variational learning framework to leverage both pseudo text and pseudo labels, and (2) incorporates two kinds of soft noise into ST, better exploring larger potential text space and extending the attribute distribution boundary. Theoretical analysis demonstrates that DuNST acts as a combination of regularization-like exploitation and attribute boundary exploration, which makes a better balance of the two requirements, significantly improving control accuracy with satisfactory generation fluency and diversity. Since the pseudo data is generated in an auto-regressive manner, which takes longer training time, we plan to explore ways to accelerate the self-training process in the future.

\section{Limitations}
Though DuNST works well, it has four kinds of limitations as follows:
\begin{itemize}
\item Decelerated training process. As with all other Self-training methods, DuNST also needs to reproduce pseudo labels and pseudo text at each ST iteration. Since the pseudo text (both hard and soft) is generated in an auto-regressive manner, which is hard to be done parallelly, leading to longer training time.

\item Reliance of unlabeled in-domain text. As we discussed in Sec.~\ref{sec:experiment}, though our soft pseudo text brings non-trivial improvement, the overall performance of all ST methods still relies on pseudo labels from unlabeled text. When unlabeled text is extremely inadequate or even unavailable (e.g., low-resource scenarios), how to better utilize pseudo text for further improvement is an open challenge.

\item Efforts of tuning noise level. As we discussed in Sec.~\ref{subsec_analysis}, the noise level $\tau$ is essential for a balanced performance, which should be carefully tuned for each downstream task.

\item Task generalization and scalability. We mainly investigate controllable NLG in this work, while it is still unknown whether our method works for other NLG tasks, like NMT and Text Summarization. Besides, as we analyzed in Sec.~\ref{subsec_theorem}, ST actually acts as a kind of regularization and smoothing. How to apply this paradigm to super large PLMs (\textit{e.g.}, GPT2-XL and GPT3), where the supervision signals from limited labeled data become extremely weak, is also an open question.
\end{itemize}

\section{Ethics Statement}
\label{sec:risk}
Since the Jigsaw dataset is unclean, the model trained on this corpus may also output some toxic and offensive expressions. Besides, our model may also be utilized to produce toxic and harmful content by simply setting the attribute label as toxic, which would take the risk of producing and propagating harmful information. Also, topic/sentiment-controlled generated text may contain some socially biased, offensive, or politically sensitive expressions. Besides, since our model significantly improves the controllability of generated text, it is likely to produce more plausible texts like fake news and movie reviews, which could possibly be utilized to produce and propagate disinformation. However, these generated texts can also be used as pseudo data in data augmentation for fake news detection and thus have the potential to increase the current fact-checking and fake news detection model. 

\section*{Acknowledgement}
Feng’s and Lakshmanan’s research was supported in part by grants from NSERC (Canada) and UBC Data Science Institute.
\bibliography{custom}
\bibliographystyle{acl_natbib}

\clearpage
\appendix
\setcounter{table}{0}
\renewcommand{\thetable}{A\arabic{table}}
\setcounter{figure}{0}
\renewcommand{\thefigure}{A\arabic{figure}}

\section{Detailed Setting}
\label{sec:appendix}
\subsection{Implementation Details}
\label{sec:impdetail}
We use pre-trained UniLM-base-cased \citep{dong2019unified} as the encoder and decoder of our VAE model since UniLM shares the parameter of transformer blocks in the encoder and decoder. We use the state of \textit{[CLS]} token to obtain the representation in the encoder. The dimension of latent $z$ is set to 128 for sentiment-controlled generation and detoxification(2-class) and 256 for topic-controlled generation (4-class). To fuse the latent $z$ better with the Transformer decoder, we use a simplified fusion method of DELLA \citep{hu-etal-2022-fuse} where we concatenate $z$ to the attention output of each token in each Transformer layer, and then add a linear layer to transfer the new attention output to the original shape of attention output. We did not use the low-rank tensor to compute layer-wise latent $z$ to save the number of parameters.

To avoid KL-vanishing, we utilize cyclical annealing tricks \citep{fu-etal-2019-cyclical} to train DuNST and set the cycle length equal to training steps in each epoch. In each cycle, first the KL weight increases from 0 to 1 linearly for the first 80\% steps in a cycle, and keeps to be 1 for the remaining 20\% steps. KL annealing is activated for 5 epochs for classification KL-loss and 7 epochs for generation KL-loss. Besides, we use the KL thresholding scheme \citep{li-etal-2020-optimus} to give up driving down KL for dimensions of $z$ that are already beneath the target compression rate \textit{KL-lambda}. 

We tuned $\textit{KL-lambda}\in\{0.01, 0.03, 0.05, 0.1\}$ (following \citet{li-etal-2020-optimus}), $\lambda_c\in\{1,5,10\}$, the ratio of Pseudo Texts (Fig. \ref{fig:numpt}), and softmax temperature $\tau\in\{0.2, 1, 5, 10\}$ (Fig. \ref{fig:temperature}) to obtain the reported results. We set \textit{KL-lambda} to be 0.05 for sentiment-controlled generation, 0.03 for detoxification, and 0.01 for topic-controlled generation. $\lambda_c$ is 10 for sentiment-controlled generation and 1 for topic-controlled generation and detoxification. Softmax temperature $\tau$ is set to be 5 for all tasks.
For other hyperparameters, $\lambda_g$ is set to be 1, and weight for BOW loss \citep{wang2017diverse} $\lambda_{bow}$ is set to be 0.2 for all tasks. We use AdamW \citep{Loshchilov2019DecoupledWD} as an optimizer. The training batch size is 8 and the learning rate is $5e-5$. We apply linear warmup to the optimizer and the number of warm-up steps is one epoch. 

We implement DuNST and all other baselines based on Huggingface Transformers \citep{wolf-etal-2020-transformers} library of v4.21.1 and use NVIDIA A100 to train our model. The total number of training GPU hours is around 8h for IMDb, 10h for Jigsaw, and 9h for AGNews. The number of parameters of our model is 134.56M for sentiment-controlled generation and text detoxification. For a topic-controlled generation, the number of parameters is 136.19M. In the generation phase, we use top-$p$ sampling ($p=0.9$) as the decoding method. Other generator configurations include a length penalty to be 1.0, a repetition penalty to be 1.0, and a no-repeat-ngram-size to be 4 for all baselines. 
All experimental results are trained and tested in a single run.

\subsection{Dataset Description}
\label{sec:datasetdesc}
For IMDb\footnote{\url{https://huggingface.co/datasets/imdb}} dataset \citep{maas-etal-2011-learning}, the authors claimed in their paper that \textit{In the interest of providing a benchmark for future work in this area, we release this dataset to the public without claiming any further copyright}.
For AGNews \footnote{\url{https://www.kaggle.com/amananandrai/ag-news-classification-dataset}} dataset \citep{Zhang2015CharacterlevelCN}, it is claimed in the website that \textit{You are encouraged to download this corpus for any non-commercial use}. 
For Jigsaw \footnote{\url{https://www.kaggle.com/c/jigsaw-toxic-comment-classification-challenge/}} dataset, the dataset is under CC0, with the underlying comment text being governed by Wikipedia's CC-SA-3.0. All datasets we used are open-sourced and are used for research only, which is consistent with their intended use.

For IMDb dataset and AGNews dataset, we leave 10\% of the training set as validation data, and others as training data. For the AGNews dataset, we use the description for text generation and wrote a script to resolve HTML tags. For Jigsaw dataset, we apply a binary setting where we keep the ``non-toxic'' class unchanged and group all other classes into ``toxic'' class.

The details of datasets are described in Table \ref{tab:dataset}. For the Jigsaw dataset, there are only 414 toxic data (9.6\%) in the Jigsaw dataset, which shows that Jigsaw is an extremely imbalanced dataset, bringing difficulty in detoxification.

\begin{table*}[ht]
 \centering
 \begin{tabular}{rccccc}\toprule
      & labeled & Unlabeled & Dev & Test &Avarage Length\\\midrule
    IMDb(5\%) & 1125 &  33750 & 2500 & 25000& 270\\
    AGNews(3\%) & 3240 & 97200 & 12000 & 7600& 41\\
    Jigsaw(3\%) & 4308 & 43080 & 15957 & 63978& 73\\
    \bottomrule
 \end{tabular}
 \caption{Description of datasets used in the experiment }
 \label{tab:dataset}
\end{table*}

\begin{table}[htbp]
 \centering
 \begin{tabular}{rccl}\toprule
              &Acc. $\uparrow$ & F1 $\uparrow$ &AUC $\uparrow$  \\\midrule
    \textit{IMDb}\\
    RoBERTa-large&96.15&96.20&99.22\\
    BERT-base&88.40&88.62&95.21\\
    \midrule
    \textit{AGNews}\\
    RoBERTa-large&94.88&94.89&99.34\\
    BERT-base&89.93&89.91&98.23\\
    \bottomrule
 \end{tabular}
 \caption{Classifier performance of our evaluator RoBERTa-large and pseudo labeler BERT-base on the test set.}
 \label{tab:eval_classifier}
\end{table}

\subsection{Evaluation Metric Details}
\label{sec:apdix-metric}
We set the minimum generation length to 10. For the maximum length, 490 for sentiment, 50 for detoxification, and 40 for topic. We evaluate generation quality on the following metrics:

\paragraph{Fluency:} We evaluate generation fluency by the perplexity of generated text measured by GPT2-XL~\citep{radford2019language}, \textit{i.e.}, \textbf{Output PPL}. 

\paragraph{Generalizability:} We calculate the perplexity of each model on each testing set, \textit{i.e.}, \textbf{Model PPL}, to evaluate the generalizability of the model. For VAE-based models, we can only obtain the lower bound of $\log p(x)$. Following \citep{li-etal-2020-optimus, hu-etal-2022-fuse}, We consider $k$ latent variables $z_1, z_2, ... , z_k$ sampled from the posterior distribution $q(z_i|x)$, and PPL based on these latents $p(x,z_i)$. Based on the fact that average importance weights are an unbiased estimator of $\log p(x)$ \citep{Burda2016ImportanceWA} and Jensen’s Inequality, we have:
\begin{equation}
\begin{aligned}
    L_k=&\mathbb{E}\left[\log \frac{1}{k} \sum_{i=1}^k \frac{p(x,z_i)}{q(z_i|x)}\right] \\
    \le& \log \mathbb{E}\left[\frac{1}{k} \sum_{i=1}^k \frac{p(x,z_i)}{q(z_i|x)}\right] = \log p(x)
\end{aligned}
\end{equation}
Thus we use $L_k$ to estimate the output PPL in VAE-like models.

\paragraph{Controllability:} We evaluate the control accuracy through classification performance (accuracy (\textbf{Acc}) and Macro-F1(\textbf{F1}) ) on the generated text by the two fine-tuned RoBERTa-large classifiers for sentiment and topic. Table \ref{tab:eval_classifier} presents the performance of our evaluator RoBERTa-large. We find that RoBERTa-large has a satisfactory classification accuracy and F1 on these two tasks, and thus is able to act as a good evaluator of generation quality. For detoxification, we report the percentage of toxic sentences (\textbf{Toxic \%}) using Google Perspective API. Perspective API is a free API for scoring the toxicity of text. Following \citet{qian-etal-2022-controllable} we also use this Perspective API for toxicity evaluation.

\paragraph{Diversity:} To evaluate the diversity of generated text, we consider the following metrics: (1) \textbf{Dist-n}~\citep{li-etal-2016-diversity}: the percentage of distinct n-grams on generated samples. We evaluate on $n=1,2,3,4$ and compute the geometric mean as \textbf{Dist}. (2) \textbf{Self-BLEU}~\citep{zhu2018texygen}. Self-Bleu calculates the BLEU score on the generated samples, which averages the BLEU score of each generated sequence calculated with other generated ones as references. The BLEU score is computed as the geometric mean of BLEU-$n$ ($n=2,3,4$). This metric measures the diversity of a set of generated sequences. Lower Self-BLEU means these generated sequences are more distinguishable from each other.

Among all the above metrics, Accuracy, F1, AUC, Dist-n, and Self-BLEU are reported as 100 times their original value for convenience.

\subsection{Baseline Details}
\label{sec:apdix-baseline}
For Finetune LM, we feed a prepend sentence as a control sentence. For sentiment-controlled generation, we use \textit{This is a [positive/negative] review} as control sentence. For topic-controlled generations, we use \textit{The following is about [topic]}. For detoxification, we use \textit{This is a [toxic/non-toxic] comment} as control sentence. For T5, since it acts in a sequence-to-sequence manner, we feed the control sentence to the encoder and training text to the decoder. We fine-tune all pre-trained LMs under learning rate 5e-5 for 10 epochs and warmup steps to be 1 epoch.

For GPT2+PT+Noise, we use the same implementation of Noise Layer as \citet{He2020Revisiting}. We set the token drop rate and mask rate to 5\%. Since GPT2 does not have a \textit{Mask} token, we randomly substitute this token for another token. We set the parameter of word shuffle to 1.1. 

For the pseudo-labeling-based method, we report the performance of pseudo labeler BERT-base-cased in Table \ref{tab:eval_classifier}. 

For GPT2+PT(select)+PL, we over-generate two times of pseudo text and compute the uncertainty score and classification confidence from the BERT-base-cased classifier. The classification confidence $s_{conf}$ is the softmax probability of the predicted label. Uncertainty score $s_{uncertain}$ is Bayesian Active Learning by Disagreement (BALD) computed by Monte-Carlo Dropout \citep{mukherjee-awadallah-2020-ust}. A high BALD score means the model is highly confused. We want to select the sample that is of high confidence and low BALD score. Thus we select samples based on the following score:
$$s_{select}=s_{conf}+\frac{1e-5}{s_{uncertain}}$$

For prefix-tuning (PF) and contrastive prefix-tuning (Ctr-PF). We follow the implementation details described in \citet{qian-etal-2022-controllable}. 

\subsection{Additional Settings for Detoxification Tasks}
\label{sec:apdix-toxic-setting}
As mentioned in \ref{sec:datasetdesc}, the Jigsaw dataset suffers from severe imbalanced labels where toxic data only counts for 9.6\% of training data. To alleviate this problem, we can tune the ratio of toxic and non-toxic data when generating pseudo texts and in conclusion balance the whole training set. We can obtain a less imbalanced dataset if we increase the ratio of toxic to non-toxic data in PT. We propose DuNST(pos) where all pseudo texts are generated from toxic attributes.

Similarly, in the baseline for detoxification tasks, we additionally tested a new variant for GPT2-based self-training methods. GPT2+PT(select, all toxic)+PL refers to all pseudo texts generated from toxic attributes, while GPT2+PT(select)+PL refers to generating 1:1 toxic/non-toxic pseudo texts.  

\subsection{Human Evaluation Details}
\label{sec:apdix-human}
We invite 5 competent annotators to score generated samples. All of the annotators are graduate students. Each annotator is assigned 420 generated text: 120/180/120 for sentiment/topic/detoxification tasks. Each sentiment/topic-controlled generated text is annotated by 2 annotators. Each text in the detoxification task is annotated by 3 annotators. 
Annotators will be paid 30\$ for completion, and they were aware of the potential risks or ethical concerns of machine-generated texts. Below we attach the instruction for workers. 

Attention: all machine-generated text involved is used for research only. The generation may contain toxic content including discrimination and offensive content.

Metrics:

Sentiment/Topic:

Fluency: Whether the generated text is fluent/grammatically correct. A small number of HTML tags/special words (like <br>) are permitted. The score ranges from 1 (least fluent) to 5 (most fluent).

Novelty: The extent of novelty INSIDE the generated text.  E.g., distinct words/expressions, and repeated phrases. The score ranges from 1 (least novel, a sentence/word repeat till the end) to 5 (most novel, all sentence describes new stuff). 

Attribute Relevance: whether the generated text is relevant to its label. There might exist examples that contain both positive/negative expressions. 
5=exactly relevant; 4=mostly relevant; 3=neutral (not contain sentiment or 50/50); 2=slightly relevant, but more similar to another class;1=totally opposite. 

Detoxification: Please only evaluate the quality of generated text AFTER prompts, since the prompt itself might be toxic.
Toxicity: whether the provided textual sequences contain any toxic content, including but not limited to offensive text, abusive language, swearwords, hate speech, denigrating messages, microaggression, discrimination, sex, rude words, and hominem attack. The score ranges from 1 (most non-toxic) to 5 (most toxic).

\section{Derivation and Proof}
\setcounter{table}{0}
\renewcommand{\thetable}{B\arabic{table}}
\setcounter{figure}{0}
\renewcommand{\thefigure}{B\arabic{figure}}

\label{sec:apdix-theorem}
\subsection{Derivation of Dual VAE ELBO}
To optimize the attribute-controllable generation direction, we aim at learning the conditional distribution of text, namely $q(x|y)$, and derive the evidence lower bound (ELBO) as:
\begin{align}
& \log q(x|y) \notag \\
 = & \log \int q(x,z|y) \frac{p(z|x,y)}{p(z|x,y)} dz \notag\\
 = &\log \mathbb{E}_{p(z|x,y)}[\frac{q(x,z|y)}{p(z|x,y)}] \notag\\
 \geq & \mathbb{E}_{p(z|x,y)}[\log \frac{q(x|z,y)q(z|y))}{p(z|x,y)}]\notag\\
 =& \mathbb{E}_{p(z|x,y)}[\log q(x|z,y)] - \text{KL}[p(z|x,y)||q(z|y)] \notag\\ 
 =& -\mathcal{L}_g,\notag 
\end{align}
where we approximate the true prior and posterior distributions $q(z|y)$, $p(z|x,y)$ with a prior network and a posterior network (a.k.a. recognition network). When we input a prompt $c$ as in our experiments on IMDB, similarly we can get $\log q(x|y,c)\geq  \mathbb{E}_{p(z|x,y,c)}[\log q(x|z,y,c)] - \text{KL}[p(z|x,y,c)||q(z|y,c)]$.

For attribute label classification, we maximize $q(y|x)$ and get a ELBO symmetrically:
\begin{align}
& \log q(y|x) \notag\\
 = &\log \int q(y,z|x) \frac{p(z|x,y)}{p(z|x,y)} dz \notag\\
= & \log \mathbb{E}_{p(z|x,y)}[\frac{q(y,z|x)}{p(z|x,y)}] \notag\\
 \geq &\mathbb{E}_{p(z|x,y)}[\log \frac{q(y|z,x)q(z|x))}{p(z|x,y)}]\notag\\
 = &\mathbb{E}_{p(z|x,y)}[\log q(y|z,y)] - \text{KL}[p(z|x,y)||q(z|x)] \notag \\
 =& -\mathcal{L}_c,\notag
\end{align}
where we similarly approximate the true prior $q(z|x)$ with another prior network.

Please note that the two optimization directions shared most parameters, and utilize the same recognition network but incorporate different prior distributions.
\subsection{Proof of Theorem 1}
For brevity, we ignore the hyper-parameters $\lambda$. Define $p$ as the real data distribution while $q$ as the estimated one. We assume we could approximate the real prior distribution of label, $q(y)$, by statistics under the i.i.d. assumption, and assume our model also estimates the real text distribution, $q(x)$, well enough with a large unlabeled dataset $D_U$. That is, $\text{KL}[p(x)||q(x)]<\epsilon$ and $\text{KL}[p(y)||q(y)]<\epsilon$. Then over the whole labeled dataset $p(x,y)$ we have:
\begin{align}
& \mathcal{L}_g + \mathcal{L}_c\notag\\ 
 = & \mathbb{E}_{p(x,y)} \{ - \mathbb{E}_{p(z|x,y)}[\log q(x|z,y)\notag\\
+&\log q(y|z,x)] + \text{KL}[p(z|x,y)||q(z|y)] \notag\\
 + &\text{KL}[p(z|x,y)||q(z|x)] \} \notag\\
 = &\mathbb{E}_{p(x,y)}\{\int p(z|x,y)[\log\frac{p(z|x,y)}{q(x|y,z)q(z|y)} \notag \\
 +& \log\frac{p(z|x,y)}{q(y|x,z)q(z|x)}] dz\}. \notag
\end{align}

Then we consider the left term of the above equation and have:
\begin{align}
& \mathbb{E}_{p(x,y)}\{\int p(z|x,y)[\log\frac{p(z|x,y)}{q(x|y,z)q(z|y)} dz \notag \\
 = & \mathbb{E}_{p(x,y,z)}\{\log \frac{p(x,y,z)q(y,z)q(y)}{p(x,y)q(x,y,z)q(y,z)}\}\notag\\
 = & \text{KL}[p(x,y,z)||q(x,y,z)]+\mathbb{E}_{p(x,y)}[\log\frac{q(y)}{p(x,y)}] \notag\\
 \approx &\text{KL}[p(x,y,z)||q(x,y,z)]+H_p(x|y) \notag \\
 \geq &\text{KL}[p(x,y,z)||q(x,y,z)],  \notag 
\end{align}
where the second last step is because by assumption we have $p(y)\approx q(y)$. Similarly, for the left term, we have:
\begin{align}
&\mathbb{E}_{p(x,y)}\{\int p(z|x,y)[\log\frac{p(z|x,y)}{q(y|x,z)q(z|x)} dz\notag\\
\approx &\text{KL}[p(x,y,z)||q(x,y,z)]+H_p(y|x). \notag \\
 \geq &\text{KL}[p(x,y,z)||q(x,y,z)]. \notag
\end{align}

Combining all the results above, we conclude:
\begin{align}
& \mathcal{L}_g + \mathcal{L}_c \geq \text{KL}[p(x,y,z)||q(x,y,z)]. 
\end{align}

Then we consider the scenario of Self-training. Define the real distribution formed by the dataset as $p(x,y,z)$, the estimated distribution at the last ST iteration as $q_{\theta}^{'}(x,y,z)$ which is formed by the generated pseudo labels and text, and the one at the current ST iteration as $q_{\theta}(x,y,z)$. As discussed in Sec.~\ref{subsec_dunst}, we add noise to pseudo text to enhance exploration. Therefore, the previously learned $q_{\theta}^{'}(x,y,z)$ is  disturbed and becomes $q_{\theta}^{'}(x,y,z) + u$ where $u$ is the noise distribution. For brevity, we abbreviate these distributions as $p$, $q_{\theta}^{'}$, $q_{\theta}$ and $u$, respectively. In Self-training, we are actually fitting $q_{\theta}$ to not only $q$ but also $q_{\theta}^{'}$ and $u$. Therefore, we are minimizing an upper bound of:
\begin{align}
& \text{KL}[p+q_{\theta}^{'}+u||q_{\theta}] \notag \\
 =& \int (p+q_{\theta}^{'}+u) \log \frac{p+q_{\theta}^{'}+u}{q_{\theta}}d. \notag
\end{align}

Consider the first term:
\begin{align}
& \int p \log \frac{p+q_{\theta}^{'}+u}{q_{\theta}}d \notag \\
 =& \int p \log \frac{p}{q_{\theta}} * \frac{p+q_{\theta}^{'}+u}{p}d \notag \\
 =& \text{KL}[p||q_{\theta}] - \text{KL}[p||p+q_{\theta}^{'}+u]. 
\end{align}

Since $p$, $q_{\theta}^{'}$ and $u$ are all fixed at the current iteration, we can ignore the last term $\text{KL}[p||p+q_{\theta}^{'}+u]$. Similarly, we have that minimizing $\text{KL}[p+q_{\theta}^{'}+u||q_{\theta}]$ equals to minimizing $\text{KL}[p||q_{\theta}]+\text{KL}[q_{\theta}^{'}||q_{\theta}]+\text{KL}[u||q_{\theta}]$, concluding the proof.

\section{Additional experimental results}
\setcounter{table}{0}
\renewcommand{\thetable}{C\arabic{table}}
\setcounter{figure}{0}
\renewcommand{\thefigure}{C\arabic{figure}}

\subsection{Detoxification Results}
\label{sec:apdix-toxic}
As shown in Table \ref{tab:toxic}, our DuNST outperforms all the other baselines on controllability. DuNST outputs the least toxic text while keeping a relatively high diversity. We find that generating all toxic pseudo texts performs better than generating 1:1 toxic/non-toxic pseudo texts for GPT2 and UniLM, which shows that adding pseudo text in self-training can tackle the issue of the imbalanced dataset. The Output PPL and Model PPL of DuNST are larger than the baselines. We explain the reason as follows. Since we are choosing toxic prompts marked as ``challenging'', it means that toxic sentences would be more likely to be generated and thus have a lower PPL score. Similarly, some non-toxic continuation might get a high PPL score from the GPT2-XL model, since it is rarer to be seen and is less natural from the challenging prompt. This does not mean that generation fluency is worse. Human evaluation on detoxification tasks (see Table \ref{tab:human-toxic}) demonstrates that DuNST generation does not have a significant difference from UniLM generation in fluency and novelty. On the other hand, its toxicity level is significantly lower than the two baselines, which further demonstrates that DuNST can improve generation controllability. 

\begin{table}[ht]
 \centering
 \small
 \begin{tabular}{rccl}\toprule
         & Fluency $\uparrow$  & Novelty $\uparrow$ & Toxicity $\downarrow$\\\midrule
    DuNST(pos)  & \textbf{3.58} & 3.83  & \textbf{1.64} \\
    Ctr-PF & 3.57 & \textbf{3.88} & 2.12$^{**}$\\
    UniLM(ST) & 3.55 & 3.72  &2.40$^{**}$\\
    \bottomrule
 \end{tabular}
 \caption{Human evaluation results on detoxification. UniLM(ST) refers to the best self-training model. ``$^{*}$'' refers to $p$-value$<0.05$. ``$^{**}$'' refers to $p$-value$<0.01$.}
 \label{tab:human-toxic}
\end{table}

\begin{table*}[htbp]
 \centering
 \begin{tabular}{rccccccccl}\toprule
    & \multicolumn{3}{c}{Sentiment}&\multicolumn{3}{c}{Topic}&\multicolumn{3}{c}{Detoxification}\\
    \cmidrule(lr){2-4}\cmidrule(lr){5-7}\cmidrule(lr){8-10}
              &Acc. $\uparrow$ & F1 $\uparrow$ &AUC $\uparrow$  &Acc. $\uparrow$ & F1 $\uparrow$ &AUC $\uparrow$&Acc. $\uparrow$ & F1 $\uparrow$ &AUC $\uparrow$\\\midrule
    DuNST & 91.8&91.9&95.8&88.8&88.8&95.8&90.3&63.0&94.4\\
    -PT & 91.7&91.8&95.5& 87.7& 87.7& 96.3 &91.5&65.6&95.4\\
    -PL-PT & 89.3&89.3&95.1&86.9&86.9&94.8&90.9&63.7&94.2\\
    \midrule
    BERT-base&88.40&88.62&95.21&89.93&89.91&98.23&91.5&64.3&95.2\\
    \bottomrule
 \end{tabular}
 \caption{Classification result.}
 \label{tab:classification}
\end{table*}

\subsection{Classification Results}
\label{sec:apdix-classfication}
Table \ref{tab:classification} reports the classification performance of our model on 3 tasks. We find that self-training on pseudo-labeled data could significantly improve classification performance, and thus improve generation quality. Pseudo texts have a slight improvement in classification performance on sentiment-controlled and topic-controlled generation tasks. For the detoxification task, the classification performance drops a little. We explain the reason as follows. We assume that attribute distribution in the test set is the same as the training set. The attribute distribution of the whole training data is shifted since all pseudo texts are toxic, which makes the distribution of classifier prediction far away from the testing set.

\begin{table}[htbp]
 \centering
 \begin{tabular}{rccl}\toprule
             & Output PPL $\downarrow$  & F1 $\uparrow$ & Dist $\uparrow$\\\midrule
    DuNST & \textbf{34.73}&\textbf{93.59}&\textbf{59.42}\\
    $-$Dual & 50.26&90.33&55.83\\
    \bottomrule
 \end{tabular}
 \caption{Comparison about duality on topic generation.}
 \label{tab:dual}
\end{table}

\begin{figure}[ht]
\centering
\includegraphics[scale=0.25]{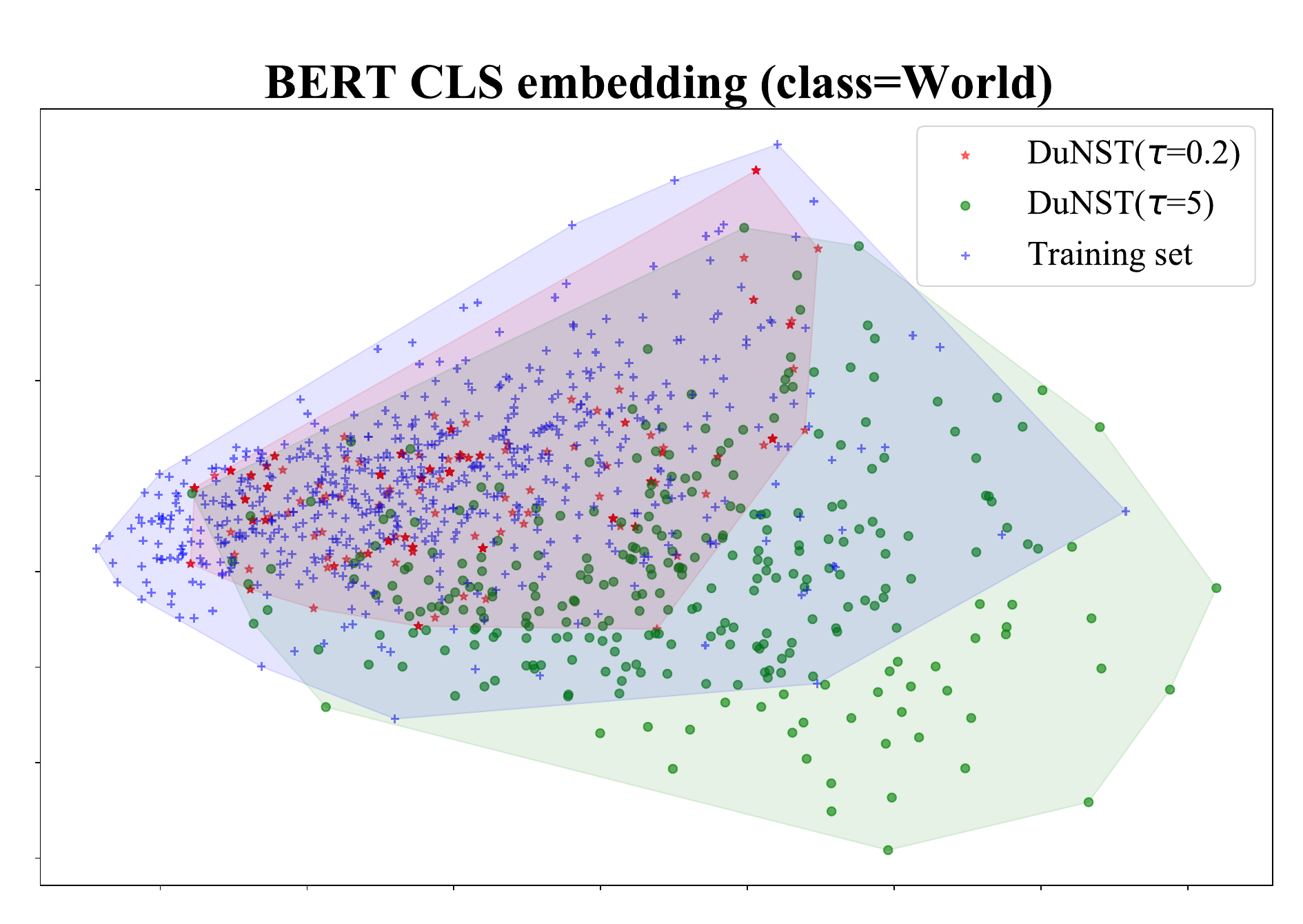}
\caption{BERT \textit{[CLS]} embedding of training texts and generated texts from DuNST model under different temperatures.}
\label{fig:bertemb}
\end{figure}

\begin{figure}[t] 

\centering 
\resizebox{0.9\columnwidth}{!}{  
			\begin{tikzpicture} 
			\scalefont{0.8} 
			\begin{axis}[
			sharp plot, 
			xmode=normal,
			xlabel=Percentage of IMDb training set, 
			ylabel=F1, 
			width=7.5cm, height=5.5cm,  
			xmin=0,xmax=101,  
			ymin=80, ymax=95,  
			xtick={5,20,50,100}, 
			ytick={80,85,90,95}, 
			xlabel near ticks, 
			ylabel near ticks, 
                ylabel style={rotate=-90},
			ymajorgrids=true, 
			grid style=dashed, 
			legend style={at={(0.8,0.02)},anchor=south}, 
			]
			\addplot+[ultra thick,mark=square,mark options={scale=0.6}, color=color1] plot coordinates {
                    (100,93.4)
                    (5,93.05)
			};
			\addlegendentry{DuNST} 
			\addplot+[ultra thick,mark=square,mark options={scale=0.6}, color=color2] plot coordinates { 
                (100,90.52)
                (50,85.91)
                (20,83.76) 
                (5,82.89)
			};
			\addlegendentry{DVAE}
   \addplot+[ultra thick, mark=x,mark options={scale=1.6}, color=color4] plot coordinates {
                    (100,90.56)
                    (50,89.57)
                    (5,88.75)
			};
			\addlegendentry{GPT2-ST} 
			\addplot+[ultra thick, mark=x, mark options={scale=1.6}, color=color3] plot coordinates {
                    (100,89.12)
                    (50,87.44)
                    (20,84.37)
                    (5,80.44)
			};
			\addlegendentry{GPT2} 
            
			\end{axis}
			\end{tikzpicture}
		}
  \caption{Generation controllability (F1) with different numbers of labeled data on IMDb dataset. Here DVAE refers to DuNST without self-training on pseudo text and pseudo data.} 
  \label{fig:numld}
\end{figure}
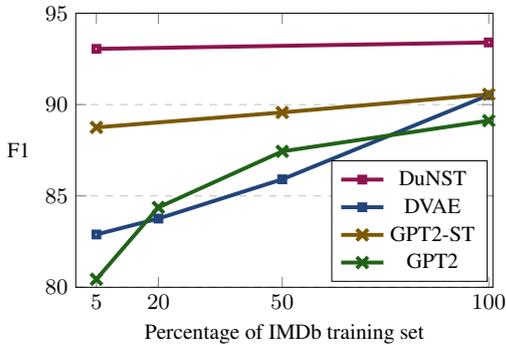

\subsection{Finer illustration of analysis}
\label{sec:apdix-analysis}
Table \ref{tab:dual} shows the comparison result on the topic generation task. We can see that without duality the generation performance drops significantly.

Fig.~\ref{fig:bertemb} depicts the distribution of BERT-large embedding of training data and DuNST-generated data in different temperatures under \textit{World} topic. Here we use the \textit{[CLS]} embedding of the BERT-large model to represent sentence embedding. We find that larger generation temperature leads to more diverse sentence representation, which demonstrates that high-temperature generation of pseudo data could improve generation diversity.

\begin{table*}[htbp]
\small
 \centering
 \begin{tabular}{lccccl}\toprule
    & \multicolumn{5}{c}{Detoxification} 
    \\
    \cmidrule(lr){2-6}
             & Output PPL $\downarrow$  & Model PPL $\downarrow$ & Toxic\% $\downarrow$ & Dist $\uparrow$ &S-BLEU $\downarrow$  \\\midrule
    Test set & 48.77& $-$ &  $-$ & 54.26 &32.22 \\
    GPT2(raw) & 25.06&10397.67&47.40&52.71&37.13\\ 
    \midrule
    \multicolumn{6}{l}{\textit{Finetune LM}} \\
    GPT2    & 32.79&66.61&43.94&51.62&42.05 \\
    UniLM & 52.23&67.92&34.38&38.26&55.31\\
    T5 & \textbf{27.21}&42.04&22.81&39.83& 63.49\\
    \midrule
    \multicolumn{6}{l}{\textit{Lightweight methods}}\\
    PF  &28.67&52.73&38.37&49.68&\textbf{41.53}\\
    Ctr-PF &29.28&57.39&31.53&49.47&46.70\\
    \midrule
    \multicolumn{6}{l}{\textit{Self-Training with GPT2}}\\
    +PT    & 36.29&71.47&41.03&\textbf{51.91}&42.15\\
    +PT+noise & 34.69&66.12&40.59&51.31&43.42\\
    +PT+PL+noise & 29.20&26.37&40.99&49.75&43.10\\
    +PT(select)+PL &29.83&26.44&43.45&49.65&43.03\\
    +PT(pos)+PL &29.49&\textbf{25.87}&40.00&49.52&43.22\\
    \midrule
    \multicolumn{6}{l}{\textit{Self-Training with UniLM}}\\
    +PT    & 46.78&74.71&34.68&36.82&55.89\\
    +PT+noise & 51.99&80.46&39.46&40.16&52.95\\
    +PT+PL+noise & 40.98&55.99&26.95&44.47&47.07\\
    +PT(select)+PL &40.70&54.50&29.21&45.42&46.94\\
    +PT(pos)+PL &45.09&55.87&25.13&45.91&46.70\\
    \midrule
    \multicolumn{6}{l}{\textit{Our Methods}}\\
    DuNST-PT & 56.75&50.28&15.32&47.03&47.10\\
    DuNST(pos) &74.74&63.75&\textbf{13.69}&50.37&42.62\\
    \bottomrule
 \end{tabular}
 \caption{Results on Jigsaw dataset. DuNST-PT refers to DuNST without pseudo text but only uses pseudo-labeled data.}
 \label{tab:toxic}
\end{table*}

\subsection{Effect of Different Numbers of Labeled Data}
We show the effectiveness of different models over changing sizes of training data in Fig. \ref{fig:numld}. We find that all models obtain improved generation performance with more labeled data, but our DuNST performs similarly when using only 5\% of labeled data compared to 100\%. Such results demonstrate the superiority of DuNST, which additionally learns from unlabeled and pseudo data through self-training. 

\subsection{Full version of experimental results}
Table \ref{tab:imdb}, Table \ref{tab:topic}, and Table \ref{tab:ablation-full} reports complete experimental results of IMDb, AGNews, and Ablation Study.

\begin{table*}[htbp]
\small
 \centering
 \begin{tabular}{lccccccl}\toprule
    & \multicolumn{7}{c}{Sentiment} 
    \\
    \cmidrule(lr){2-8}
             & Output PPL $\downarrow$  & Model PPL $\downarrow$ & Acc $\uparrow$ & F1 $\uparrow$ & AUC $\uparrow$ & Dist $\uparrow$ &S-BLEU $\downarrow$  \\\midrule
    Test set & 25.14&$-$ &96.15&96.20&99.22& 48.27 & 43.34\\
    GPT2(raw) & 13.20&38.39&55.90&68.50&61.37&35.91&58.79 \\
    \midrule
    \multicolumn{7}{l}{\textit{Finetune LM}} \\
    GPT2    &16.40&44.02&77.55&80.44&88.35&26.34&71.00 \\
    UniLM & 25.20&54.33&76.45&75.35&85.18&31.05&66.97 \\
    T5 & 25.69&34.97&82.80&83.77&90.50&30.03&69.57\\
    \midrule
    \multicolumn{7}{l}{\textit{Lightweight method}}\\
    PF  &13.02&37.09&67.55&75.05&81.84&29.48&65.10\\
    Ctr-PF &13.01&37.12&71.00&77.33&86.51&29.63&\textbf{64.83}\\
    \midrule
    \multicolumn{7}{l}{\textit{Self-Training with GPT2}}\\
    +PT    &  14.62 & 68.04&76.10&79.57&87.92&30.58&65.22 \\
    +PT+noise &11.91 &44.31&74.95&77.46&85.02&25.40&72.19 \\
    +PT(noise)+PL &11.26&33.85&87.60&88.47&95.59&27.26&70.90 \\
    +PT(select)+PL &\textbf{10.89}&33.89&88.32&88.75&96.24&27.17&71.41 \\
    \midrule
    \multicolumn{7}{l}{\textit{Self-Training with UniLM}}\\
    +PT    & 26.62 &58.37&72.2&70.27&80.37&31.17&66.69 \\
    +PT+noise &30.28&62.07&77.75&75.78&85.35&31.68&65.18 \\
    +PT(noise)+PL &18.92&\textbf{33.53}&89.95&89.73&96.38&30.94&66.84 \\
    +PT(select)+PL &18.40&33.56&90.08&90.06&96.66&31.27&67.61 \\
    \midrule
    \multicolumn{7}{l}{\textit{Our Methods}}\\
    DuNST  & 21.67&42.82&\textbf{92.90}&\textbf{93.05}&\textbf{98.02}&\textbf{31.79}&65.80\\
    \bottomrule
 \end{tabular}
 \caption{Results on IMDb dataset.}
 \label{tab:imdb}
\end{table*}

\begin{table*}[htbp]
\small
 \centering
 \begin{tabular}{lccccccl}\toprule
    & \multicolumn{7}{c}{Topic} 
    \\
    \cmidrule(lr){2-8}
             & Output PPL $\downarrow$  & Model PPL $\downarrow$ & Acc $\uparrow$ & F1 $\uparrow$ & AUC $\uparrow$ & Dist $\uparrow$  &S-BLEU $\downarrow$  \\\midrule
    Test set &31.04&$-$ & 94.88&94.89&99.34 & 67.24 & 23.31 \\
    GPT2(raw) & 16.94 & 74.41& 55.75 & 52.17 & 83.28& 46.88&45.55 \\
    \midrule
    \multicolumn{7}{l}{\textit{Finetune LM}} \\
    GPT2    &22.22 &23.46&82.92&83.08&95.23&54.93&39.93 \\
    UniLM & 55.79 &36.28&87.67&87.70&96.30&54.76&43.77 \\
    T5 & 48.33&32.12&88.33&88.43&97.95&58.06&37.01\\
    \midrule
    \multicolumn{7}{l}{\textit{Lightweight method}}\\
    PF  &\textbf{20.27}&32.35&68.67&68.44&87.14&59.17&32.73\\
    Ctr-PF &20.41 &33.90&83.25&83.21&95.47&\textbf{60.34}&\textbf{31.20}\\
    \midrule
    \multicolumn{7}{l}{\textit{Self-Training with GPT2}}\\
    +PT    & 23.74&27.88&83.50&83.55&95.49&57.89&36.02\\
    +PT+noise & 26.39&27.02&82.42&82.45&94.58&58.06&35.53\\
    +PT(noise)+PL &30.62&\textbf{13.96}&87.83&87.48&97.42&47.11&56.67 \\
    +PT(select)+PL &31.34&14.07&87.92&87.54&97.46&46.71&57.33 \\
    \midrule
    \multicolumn{7}{l}{\textit{Self-Training with UniLM}}\\
    +PT    & 57.40&40.95&86.42&86.36&96.69&52.35&46.41\\
    +PT+noise & 58.59&45.32&85.42&85.27&95.88&53.35&46.57\\
    +PT(noise)+PL &32.36&16.64&89.67&89.70&98.11&53.79&47.95 \\
    +PT(select)+PL &33.23&16.66&90.5&90.52&98.31&53.71&47.69 \\
    \midrule
    \multicolumn{7}{l}{\textit{Our Methods}}\\
    DuNST & 34.73&33.58&\textbf{93.58}&\textbf{93.59}&\textbf{98. 99}&59.42&37.02\\
    \bottomrule
 \end{tabular}
 \caption{Results on AGNews dataset.}
 \label{tab:topic}
\end{table*}

\begin{table*}[htbp]
\small
 \centering
 \begin{tabular}{rccccccl}\toprule
    & \multicolumn{7}{c}{IMDb} 
    \\
    \cmidrule(lr){2-8}
             & O-PPL $\downarrow$  &M-PPL $\downarrow$ & Acc $\uparrow$ & F1 $\uparrow$ & AUC $\uparrow$ & Dist $\uparrow$ &S-BLEU $\downarrow$  \\\midrule
    DuNST  & 21.67&42.82&\textbf{92.9}&\textbf{93.05}&\textbf{98.02}&31.79&\textbf{65.80}\\
    $-$VAE &19.67&38.56&92.11&92.12&97.85&31.39&66.21\\
    $-$SPT &\textbf{18.53}&\textbf{36.53}&91.55&91.64&96.97&31.51&67.07\\
    $-$PT&20.91&41.14&91.7&91.83&96.93&31.67&66.27\\
    $-$PL &47.45&197.27&83.0&83.41&91.48&32.61&66.17\\
    $-$PL-SPT& 48.56&219.30&81.1&80.85&90.05&\textbf{32.86}&66.12\\
    $-$PL$-$PT&42.12&147.14&82.7&82.89&91.46&29.75&68.91\\
    \bottomrule
 \end{tabular}
 \caption{Full ablation study results on IMDb dataset. PT: pseudo text. SPT: soft pseudo text. PL: pseudo label. The
symbol $-$ means removing the settings from DuNST.$-$VAE reduces to jointly learning of classifier and generator. $-$PL$-$PT reduces to the naive dual VAE.}
 \label{tab:ablation-full}
\end{table*}

\section{Example of Generation}
\setcounter{table}{0}
\renewcommand{\thetable}{D\arabic{table}}
\setcounter{figure}{0}
\renewcommand{\thefigure}{D\arabic{figure}}
\label{sec:apdix-case}
We sample some generated texts based on Ctr-PF, GPT2-ST, UniLM-ST, and DuNST and place them on Table \ref{tab:sentiment-case}, \ref{tab:sentiment-case-continue}, Table \ref{tab:topic-case} and Table \ref{tab:topic-case-continue}. Due to the offensive content generated by these models in the detoxification task, we do not include their examples in detoxification experiments.

\begin{table*}[ht]
 \centering
 \begin{tabular}{p{1.2cm}|p{14cm}}\toprule
         Model & Generation\\\midrule
    &\textbf{Sentiment}: \textit{\negative{Negative}}; \textbf{Prompt}:\textit{At 20 years old}\\
    Ctr-PF & \textit{At 20 years old}, \negative{ichorous} and \negative{unrepentant}, ichanarito is a \positive{well-liked} celebrity, but a \negative{poor quality actor}, and a \negative{poor quality movie}... \\
    GPT2-ST & \textit{At 20 years old},  I grew up with it and \negative{had no expectations}.  I was a \positive{big fan} of the series, I thought it was \positive{great}, and I wanted to see more of it.  But soon enough, I \negative{got bored} and just \negative{got bored} watching it...\\
    UniLM-ST & \textit{At 20 years old}, I \negative{didn't expect much} when I heard this movie was being made. My initial reaction was that it was an art - house movie ( a very \negative{low budget} movie at that ), but when I saw how they did it, it just got \negative{boring} and \negative{boring}...\\
    DuNST  & \textit{At 20 years old}, the \negative{poorest} hour of my life had me scratching my head. This was an \negative{insult} to the intelligence of mankind, and a complete \negative{disregard} for anyone else's intelligence. \negative{Do not watch} this movie. It is a \negative{disgrace} to any religion...\\
    \midrule
    &\textbf{Sentiment}: \textit{\positive{Positive}}; \textbf{Prompt}:\textit{At 20 years old}\\
    Ctr-PF&\textit{At 20 years old}, ixxoxx was a fan \positive{favorite and loved} his work and movies like it was his own and his own. A hard worker who enjoyed getting his hands dirty and trying to figure out the best ways to make a buck, ixixx made his name on film and television...\\
    GPT2-ST & \textit{At 20 years old}, ixTch was my first exposure to the French film industry. I immediately knew the "circles" of French cinema, where each character had a particular expression in their mind, but the film itself was \negative{only a very short one}. I was not interested in the story or the plot, just in the characters...\\
    UniLM-ST & \textit{At 20 years old}, it is still one of my \positive{favorite} films and the music still rings true. I have seen it countless times and each time I hear it, it gets \positive{better} and \positive{better}...\\
    DuNST & \textit{At 20 years old}, I was 13 when this came out. I remember it as being a typical 1950's soap opera. The writing was \positive{excellent}. It was \positive{realistic}, the characters were \positive{believable}. The story was \positive{accurate}, but the real problems the family had with adult life in particular were out to bother with... \\
    \bottomrule
 \end{tabular}
 \caption{Example text for sentiment-controlled generation. Words in \positive{blue}/\negative{red} are positive/negative indicators, respectively.}
 \label{tab:sentiment-case}
\end{table*}

\begin{table*}[ht]
 \centering
 \begin{tabular}{p{1.2cm}|p{14cm}}\toprule
         Model & Generation\\\midrule
    &\textbf{Sentiment}: \textit{\negative{Negative}}; \textbf{Prompt}:\textit{The book}\\
    Ctr-PF &\textit{The book} I \positive{loved} a lot about The Legend of Zelda: Breath of the Wild. Its protagonist, Zelda, is an ordinary woman who grows up playing on a farm. She gets to choose a young boy in her village who can only dream about Zelda. The village gets more and more out of her. She tries to save the villagers by making them \positive{happy} by getting them into a happy life. The game takes this into an extreme (it's not like Zelda is a good actress, just a \negative{poor} one)...\\
    GPT2-ST&\textit{The book}, "Fatal Attraction" by R.D.Hilbert, was a \positive{great} book and even if the book had been better, they probably would have given it a \positive{10 out of 10}. I was \negative{disappointed} by this movie. I do not believe that a movie can be this \negative{boring}, with actors playing their roles and acting like the movie would give it a realistic look...\\
    UniLM-ST & \textit{The book} was \positive{wonderful}. The film is a \negative{mess}. I could watch this and then get \negative{frustrated}.  It is very much like an extended stage play. The actors do what they can with little money and they do a \positive{wonderful} job in acting. I will not make this up.\\
    DuNST & \textit{The book} was a rather \negative{confusing} and \negative{disappointing} adaptation of a classic. It may very well have been a good book if it had been adapted to a movie, but this movie \negative{could have been much better}. It's impossible to pinpoint exactly what exactly was the point of this film...\\
    \midrule
    &\textbf{Sentiment}: \textit{\positive{Positive}}; \textbf{Prompt}:\textit{The book}\\
    Ctr-PF &\textit{The book} has an overall feel to it and is quite \positive{good}. There are plenty of laughs and a \positive{decent ending} to the book. However, there are some other \negative{issues} with the author, who has some \negative{questionable} editing that can \negative{ruin} the book, making the movie and television series \positive{too good to be true}...\\
    GPT2-ST&\textit{The book} is a perfect example of why not to make a movie. The dialogue, the actors, the sets, everything. That said, the book is \positive{perfect}, despite the fact that it has very little plot, and that is \positive{good}. The acting is \positive{good} and is also very \positive{funny}. The story is even more \positive{funny} than the book...\\
    UniLM-ST & \textit{The book} was \positive{excellent}. And it was not only about an alien invasion of Earth, but a series of events in the near future. I was a bit \negative{reluctant} to read the book as it was written after the global collapse of the european market. But then again, it would still be hard to buy the movie More or less, the book is just brilliant, with \positive{wonderful} characters and \positive{good} animation...\\
    DuNST & \textit{The book} ( which I'm in a "Truly Hip" ) is a \positive{fantastic} comic book. The movie adaptation of the book is a \positive{superb} one. It's \positive{faithful} to the spirit of the book and looks to the footage and characters. I highly \positive{recommend} this film if you like and appreciate the story... \\
    
    \bottomrule
 \end{tabular}
 \caption{Example text for sentiment-controlled generation continued. Words in \positive{blue}/\negative{red} are positive/negative indicators, respectively.}
 \label{tab:sentiment-case-continue}
\end{table*}

\begin{table*}[ht]
 \centering
 \begin{tabular}{p{1.2cm}|p{14cm}}\toprule
         Model & Generation\\\midrule
    &\textbf{Topic}: \textit{World}\\
    Ctr-PF & (1)On Wednesday, the Iraqi security forces stormed the main Islamic State office, killing nine members, killing 12 people, but did not withdraw from the building and taking over two other buildings. \par(2)He said this could not be confirmed, but was quoted by Reuters. Hide Caption : A woman reacts after her car was blown up near an Islamist group rally in Dhaka. \\
    GPT2-ST & (1)LOS ANGELES (Reuters) - A former crematory operator  agreed on Wednesday to plead guilty to dumping  bodies and ashes at the same crematory  site where he was born, officials said\par(2)GAZA (Reuters) - A rocket killed two Israeli  soldiers in Gaza on Wednesday, the first time the  army in occupied territory in more than a year that  Hamas militants have launched a fierce...\\
    UniLM-ST & (1) The Israeli Army has suspended a company commander accused of emptying an ammunition clip into a 13 - year - old Palestinian girl. \par (2) BAGHDAD, Iraq - A roadside bomb killed two American soldiers and wounded three others in Iraq, the U. S. command said Friday, as insurgents hit Baghdad targets with rocket and rocket bombs...\\
    DuNST & (1) AP - An Italian aid worker walked free from the southern Philippines on Sunday, a day after he was abducted at gunpoint on the streets of Real Aires. \par (2) AFP - The United States and South Korea failed to hammer out a deal over a timetable for the planned reduction of US forces in Iraq, with Seoul asking for more troops to join another group.\\
    \midrule
    &\textbf{Topic}: \textit{Sport}\\
    Ctr-PF & (1)This past weekend, when the Los Angeles Lakers drafted Michael Jordan, he looked like a real contender to play the role of mentor at the small forward spot. \par(2)Houston is now playing "The Voice of America" at Madison Square Garden. The Knicks are 0-5 and facing a 10-point Los Angeles Lakers team that, if they win tonight...\\
    GPT2-ST & (1) SEATTLE (Reuters) - Olympic chiefs may have to  reconsider their decision to stage a one-day event  in Atlanta after protests from marchers in the  southern city... \par(2)NEW YORK (Reuters) - Tommy Haas looked as though he had  the flu, as he sat in his BMW 712 at  the World Championship in Akron, Ohio, on Friday.\\
    UniLM-ST & (1) England coach Sven - Goran Eriksson says striker Michael Owen must prove in training Monday that he deserves to face Wales in a World Cup qualifier. \par (2) ATHENS - - The tears were from the Brazilian women's soccer team, who had just won their first Olympic gold medal in women's tennis, beating Australia 4 - 2 in the final... \\
    DuNST & (1) South Carolina assistant Skip Holtz left the game with an injured tailback Ciatrick Fason. Freshman Adrian Peterson rushed for 140 yards and two touchdowns and Ronnie Brown added 127 yards. \par(2) AP - The New York Yankees wasted little time getting down to business, and their starting pitcher, Tony Womack, was allowed to sit out Saturday night after missing two games because of an elbow injury... \\
    \bottomrule
 \end{tabular}
 \caption{Example text for topic-controlled generation.}
 \label{tab:topic-case}
\end{table*}

\begin{table*}[ht]
 \centering
 \begin{tabular}{p{1.2cm}|p{14cm}}\toprule
         Model & Generation\\\midrule
    &\textbf{Topic}: \textit{Business}\\
    Ctr-PF& (1)President Barack Obama has said his administration is "very concerned about Iran's nuclear program and concerns about the growing threat from terrorist groups in Iran. \par(2) A number of firms have taken steps to make their online business more efficient and more efficient. New York-based Gartner says that new companies such as AT\&T, Bell, IBM...\\
    GPT2-ST& (1)NEW YORK (Reuters) - U.S. blue chips sank on Thursday  after Ford Motor Co.\par(2)SINGAPORE (Reuters) - Asian stock markets opened lower on  Thursday, helped by poor weather forecasts and gains by  technology firms, but some oil-related stocks remained higher.\\
    UniLM-ST & (1) TORONTO ( CP ) - Stock markets were poised for an early rally Thursday as crude oil prices reached record highs and energy stocks surged on easing supply fears.\par (2) In the latest move by the US Justice Department, The Washington Post has announced that it will pay \$ 60 million cash to buy the parent company of CBS MarketWatch.\\
    DuNST & (1) Tokyo stocks plunged Monday morning as investors took profits from recent gains. The US dollar was up against the Japanese yen. The Nikkei Stock Average of 225 issues was up 36.  \par(2) NEW YORK, Aug 18 ( Reuters ) - Rupert Murdoch's News Corp. Ltd. has agreed to sell its stake in Sky Latin America to DirecTV Group D \\
    \midrule
    &\textbf{Topic}: \textit{Sci/Tech}\\
    Ctr-PF &(1)\$1,000 for 'Millionaire's 'Rape Crisis' Victim Fund
By Michael S. Osterholm - 6/9/17 07:08:04: \par(2) US government has approved a \$10 million loan to provide medical equipment to the Palestinian Authority for 'humanitarian and medical equipment' on the West Bank. The agreement provides \$3 million for a...\\
    GPT2-ST& (1)NEW YORK (Reuters) - The U.S. Securities and Exchange  Commission has voted 5-0 to recommend that Internet advertising services stop soliciting fees from Web sites,  according to a...\par(2)LOS ANGELES (Reuters) - "Cell " phones offer fast data  rates, low prices and no worries about getting fat in the  long run, says a survey by analysts at research firm...\\
    UniLM-ST & (1) Sony Corp.'s music unit is abandoning its CDs that use built - in technology that limits copying them, after pushing the program for two years.\par (2) The future of the internet could be in doubt in around two years'time, according to two leading internet watchers, who outlined a series of steps they hope will turn the internet into a business...\\
    DuNST & (1) Toshiba has announced a new transmission system for routers and switches that will improve automatic transmission rates. The 6500 Super\_GSM/GPRS system will feature high clock speed...  \par(2) At a press conference this week, Bill Murray, Microsoft's CEO, expressed doubt that the software giant's strategy for regaining PC identity is considerable, but heard little reason to believe it \\
    \bottomrule
 \end{tabular}
 \caption{Example text for topic-controlled generation (continued).}
 \label{tab:topic-case-continue}
\end{table*}

\end{document}